\documentclass[10pt, journal, compsoc]{IEEEtran}
\usepackage{newtxtext,newtxmath}
\usepackage[utf8]{inputenc}
\usepackage{booktabs}
\DeclareUnicodeCharacter{2212}{\textminus}
\usepackage{amsmath, amsfonts}
\usepackage{amsmath}
\usepackage{bbm}
\usepackage{algorithmic}
\usepackage{algorithm}
\usepackage{array}
\usepackage[caption=false, font=footnotesize, labelfont=sf, textfont=sf]{subfig}
\usepackage{textcomp}
\usepackage{stfloats}
\usepackage{url}
\usepackage{verbatim}
\usepackage{graphicx}
\usepackage{multirow} %导言区
\usepackage{booktabs}
\usepackage{graphicx, subfig}
\usepackage{cite}
\usepackage{verbatim}
\usepackage{epstopdf}
\usepackage[section]{placeins}
\usepackage{float}
\usepackage{hyperref}
\usepackage[capitalize]{cleveref}
\usepackage{xcolor}
\usepackage{newtxtext, newtxmath}
\usepackage{hyperref}
\usepackage{url}
\usepackage[table]{xcolor}
\usepackage{booktabs}
\usepackage[compatibility=false]{caption}
\usepackage{bm}
\usepackage{ragged2e}
\usepackage{bbding}

\newcommand{\best}[1]{\cellcolor{blue!25}\textbf{#1}}
\newcommand{\second}[1]{\cellcolor{blue!10}#1}

\hypersetup{
colorlinks=true, % 激活链接颜色，去掉链接边框
linkcolor=red, % 文档内部链接颜色（如图表等引用）
citecolor=green, % 文献引用链接颜色
filecolor=mycustompurple, % 文件链接颜色
urlcolor=magenta % 外部URL链接颜色
}

\begin{document}
%
% paper title
\title{Vector Map as Language: Toward Unified Remote Sensing Vector Mapping}

\author{Yinglong~Yan, Yunkai~Yang, Haoyi~Wang, Wei~Fu, Linshan~Wu, Honghu~Pan, Shaobo~Xia, Shanghang~Zhang, Hao~Chen,~\IEEEmembership{Senior~Member,~IEEE},~and~Leyuan~Fang,~\IEEEmembership{Senior~Member,~IEEE}

\IEEEcompsocitemizethanks{ \IEEEcompsocthanksitem Yinglong Yan, Yunkai~Yang, Haoyi~Wang, Wei~Fu, Honghu Pan and Leyuan Fang are with the school of Artificial Intelligence and Robotics, Hunan University, Changsha 410082, China. E-mail: \{yanyl, yunkai, wanghaoyi\}@hnu.edu.cn, fuweiii1024@outlook.com, honghupan@hnu.edu.cn, fangleyuan@gmail.com.

\IEEEcompsocthanksitem Linshan Wu and Hao Chen are with the Department of Computer Science and Engineering, The Hong Kong University of Science and Technology, Clear Water Bay, Hong Kong. E-mail: linshan.wu@connect.ust.hk, jhc@cse.ust.hk.

\IEEEcompsocthanksitem Shaobo Xia is in the Department of Geomatics Engineering, Changsha University of Science and Technology, Changsha 410114, China. E-mail: shaobo.xia@csust.edu.cn.

\IEEEcompsocthanksitem Shanghang Zhang is in State Key Laboratory of Multimedia Information Processing, School of Computer Science, Peking University, E-mail: shanghang@pku.edu.cn.

\IEEEcompsocthanksitem Corresponding author: Leyuan Fang}% <-this % stops an unwanted space
}
% The paper headers
\markboth{Journal of \LaTeX\ Class Files,~2026}%
{Shell \MakeLowercase{\textit{et al.}}: Vector Map as Language: Toward Unified Remote Sensing Vector Mapping}

% in the abstract or keywords.
\IEEEtitleabstractindextext{%
\begin{abstract}
\justifying
Remote sensing vector mapping aims to generate structured maps of geospatial entities, such as buildings, roads, and water bodies, from remote sensing imagery. In practice, vector maps usually contain multiple category layers and heterogeneous entity structures, requiring a unified model for diverse mapping needs. However, existing methods typically represent vector objects as polygons or graphs, making them suitable only for specific categories: polygons poorly capture topological relations, while graphs often blur instance boundaries. We observe that language, as a natural medium for human communication, offers a flexible and expressive representation that can accommodate heterogeneous map elements, including geometry, semantics, and topolog. Motivated by this insight, we propose Vector Map as Language (VecLang), a unified paradigm that reformulates multiclass vector mapping as structured text generation. VecLang encodes the common elements of different geospatial entities into a GeoJSON-like vector language, enabling cross-category modeling within a shared textual format. To generate this language reliably, we design a progressive vision-language mapping framework that first localizes vectorization units and then generates structured map elements. We further introduce Hierarchical Vector Language Optimization, which uses reinforcement learning to improve syntax validity, content fidelity, and map executability. We also build VecMap-Bench with 54K images and 800K instances, supporting training and evaluation across standard and generalization settings. Extensive experiments demonstrate that VecLang handles both single-class and multiclass vector mapping while achieving strong cross-dataset and open-vocabulary generalization. The model and dataset are publicly available at \href{https://github.com/yyyyll0ss/VecLang}{https://github.com/yyyyll0ss/VecLang}.
\end{abstract}

% Note that keywords are not normally used for peerreview papers.
}

% make the title area
\maketitle

\IEEEdisplaynontitleabstractindextext

\IEEEpeerreviewmaketitle

\IEEEraisesectionheading{\section{Introduction}\label{sec:introduction}}

\IEEEPARstart{R}{emote} sensing vector mapping (RSVM) aims to extract the geometric boundaries, 
semantic categories, and topological relations of different geospatial entities from remote sensing imagery, 
and is an important task in remote sensing visual understanding~\cite{li2026full, wang2023learning, zhu2017deep}. 
Compared with raster representations, vector maps are typically more compact and efficient, and are better suited for editing, storage, and spatial analysis~\cite{longley2015geographic, girard2021polygonal, yan2023vectorization}. 
These advantages make them highly valuable for applications such as urban planning, traffic management, 
and environmental monitoring~\cite{lacoste2005point, liao2025maptrv2, zheng2023farsegplusplus, liu2025generating}. 
With the rapid growth of high-resolution remote sensing imagery, automatically generating high-quality, multiclass vector maps from complex scenes has become a key problem in remote sensing visual understanding~\cite{vanetten2018spacenet,girard2021polygonal,he2020sat2graph}.

\begin{figure}[!t]
    \centering
    \includegraphics[width=0.5\textwidth]{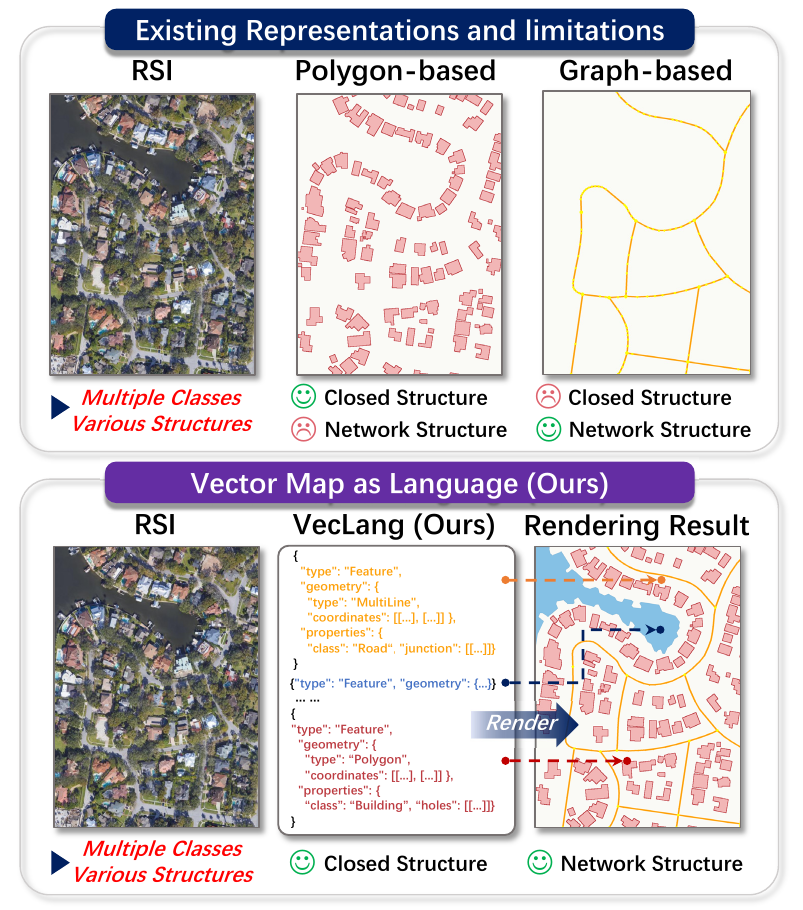}
    \caption{Comparison between existing methods \cite{jiao2026acpvnet, hetang2024samroad} and VecLang. VecLang represents diverse geospatial entities as language, enabling unified modeling of closed and network-like structures.}
    \label{fig_first_show}
\end{figure}

In practice, vector maps usually contain multiple category layers and heterogeneous entity structures, requiring a unified model to accommodate diverse mapping needs. Although RSVM has advanced rapidly in recent years, existing methods remain largely tailored to specific categories or tasks \cite{li2019topological, yan2025univector, jiao2026acpvnet}, and thus lack a unified paradigm for multiclass modeling. This limitation largely stems from the choice of representation, which determines whether different geospatial entities can be described and reasoned over within a single framework. As shown in Fig.~\ref{fig_first_show}, existing methods mainly fall into two groups: polygon-based and graph-based representations. Polygon-based methods \cite{girard2021polygonal, wei2023hisup, zhang2024p2pformer, zhang2026vectorllm} usually model vector objects as sequences of polygon vertices and recover object outlines through coordinate regression and mask constraints. They are well suited for closed objects with clear boundaries, such as buildings, but struggle to naturally represent topological connections in network-like objects, e.g., roads. Graph-based methods \cite{he2020sat2graph, hetang2024samroad, yin2025samroadplusplus}, by contrast, represent vector objects as graphs composed of nodes and edges. While they better capture relational information, they often have difficulty clearly distinguishing independent object instances. As a result, both types of representations struggle to jointly model geometric, semantic, and topological information across multiple categories within a single framework \cite{yan2025univector,jiao2026acpvnet}. In summary, representation limitations have become a key bottleneck for unified modeling and generalization in multiclass remote sensing vector mapping.

\begin{figure*}[!t]
    \centering
    \includegraphics[width=1.0\textwidth]{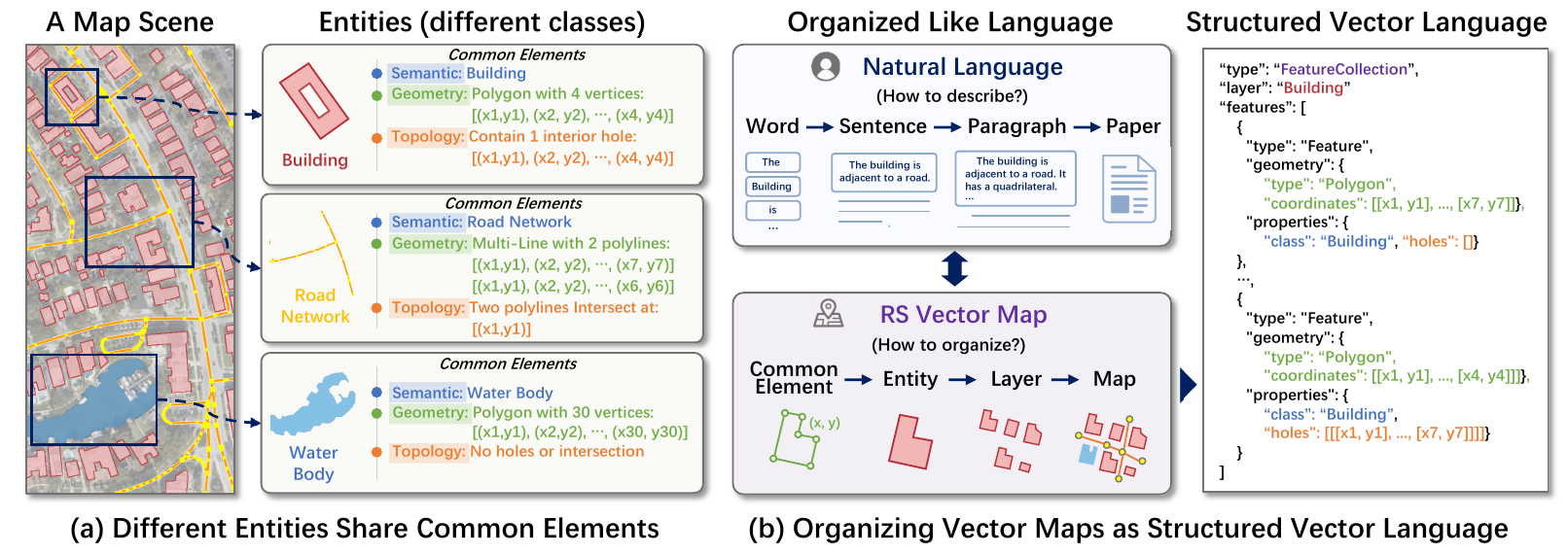}
    \caption{Motivation of our work. Different map entities share common elements, including semantics, geometry, and topology. Inspired by the organization of natural language, we structure these elements into a hierarchical vector language, enabling unified modeling of multiple vector mapping tasks.}
    \label{fig_motivation}
\end{figure*}

This limitation calls for a unified and extensible representation for heterogeneous geospatial entities. Although buildings, roads, and water bodies differ in geometry and spatial organization, their vector-map descriptions share three essential elements: geometry, semantics, and topology, as illustrated in Fig.~\ref{fig_motivation}(a). This observation motivates us to organize multiclass vector mapping based on common map elements rather than category-specific output forms. As a natural medium for human communication, language inherently provides a unified, hierarchical, and extensible way to organize information, making it well suited for describing diverse map elements and their relations~\cite{fedorenko2024language,lake2023humanlike}. As illustrated in Fig.~\ref{fig_motivation}(b), we therefore explore representing vector maps as a structured language, in which different entities follow a common grammar while preserving their category-specific geometry and topology. Based on this formulation, we propose Vector Map as Language (VecLang), a unified paradigm that reformulates multiclass vector mapping as structured text generation. By encoding diverse geospatial entities in a shared language space, VecLang enables unified modeling and cross-category generalization.

Following this idea, we develop VecLang from three aspects: representation, generation, and optimization. First, we propose Structured Vector Language (SVL), a GeoJSON-like language~\cite{butler2016geojson} that unifies geometry, semantics, and topology in a GIS-compatible format. SVL uses the same structured grammar to describe both closed objects, such as buildings and water bodies, and network-like objects, e.g., roads. Second, we design a vision-language-model-based Progressive Vectorization Framework (PVF), which first localizes vectorization units and then generates structured map elements for each unit. This progressive decomposition turns scene-level dense vectorization into localized unit-level generation, reducing visual complexity and sequence length. Third, we introduce Hierarchical Vector Language Optimization (HVLO), which uses hierarchical rewards to optimize SVL generation from syntax, content, and execution levels, bridging text generation with executable vector maps.

To train and evaluate VecLang, we construct \textbf{VecMap-Bench}, a comprehensive benchmark with approximately 54K images and 800K instances from multiple public datasets \cite{ji2019whu,he2020sat2graph,zhang2017evlabss,tong2020gid,zhang2023aqsnet,meng2025irsamap,mohanty2020deep,vanetten2018spacenet,lin2014coco,zamir2019isaid}. As shown in Fig.~\ref{fig_benchmark}, it covers single-class, multiclass, cross-dataset, and open-vocabulary settings. All annotations are cleaned, simplified, and converted into SVL through a unified map-to-language pipeline. Experiments show that VecLang achieves strong geometric accuracy, unified multiclass modeling ability, and robust cross-dataset and open-vocabulary generalization. Experimental results demonstrate that our VecLang can jointly handle polygonal objects and network-like structures within a single framework, validating structured language as an effective representation for vector maps. Our main contributions can be summarized as follows:
\begin{itemize}
    \item {\textbf{Paradigm:} To unify the modeling of different categories in vector mapping, we propose Vector Map as Language (VecLang), a new paradigm that represents vector maps with Structured Vector Language (SVL) and reformulates remote sensing vector mapping as structured text generation.}

    \item {\textbf{Framework:} To generate SVL reliably in dense remote sensing scenes, we develop a complete VecLang framework with Progressive Vectorization Framework (PVF) and Hierarchical Vector Language Optimization (HVLO), enabling stable generation and executable vector maps.}

    \item {\textbf{Benchmark:} To evaluate unified vector mapping and generalization, we construct VecMap-Bench, a benchmark with about 54K images and 800K instances across single-class, multiclass, cross-dataset, and open-vocabulary settings.}
\end{itemize}

\section{Related Work}
\subsection{Remote Sensing Vector Mapping}
Remote sensing vector mapping aims to recover explicit geometric structures of geospatial entities from remote sensing imagery, and serves as an important foundation for geographic information analysis and map updating. Existing methods are mostly designed for specific categories or tasks, and can be broadly divided into polygon-based methods \cite{wei2023hisup, zhang2024p2pformer, zhang2026vectorllm} and graph-based methods \cite{he2020sat2graph, hetang2024samroad, yin2025samroadplusplus}.

\begin{table}[!t]
\centering
\caption{Comparison of vector mapping methods. ``Closed'' and ``Network'' indicate whether the method models closed objects and network-like structures, respectively. ``Unified'' indicates whether a single model can be applied to multiple vector mapping tasks.}
\label{tab:method_summary}
\resizebox{\columnwidth}{!}{
\begin{tabular}{lcccc}
\toprule
Method & Pub. & Closed & Network & Unified \\
\midrule
HiSup \cite{wei2023hisup} & ISPRS23 & \CheckmarkBold & \XSolidBrush & \XSolidBrush \\
P2PFormer \cite{zhang2024p2pformer} & TGRS24 & \CheckmarkBold & \XSolidBrush & \XSolidBrush \\
HoliTracer \cite{wang2025holitracer} & ICCV25 & \CheckmarkBold & \XSolidBrush & \XSolidBrush \\
VectorLLM \cite{zhang2026vectorllm} & ISPRS26 & \CheckmarkBold & \XSolidBrush & \XSolidBrush \\
PFNet \& IPNet \cite{li2026full} & TPAMI26 & \CheckmarkBold & \XSolidBrush & \XSolidBrush \\
ACPV-Net \cite{jiao2026acpvnet} & CVPR26 & \CheckmarkBold & \XSolidBrush & \XSolidBrush \\
\midrule
Sat2Graph \cite{he2020sat2graph} & ECCV20 & \XSolidBrush & \CheckmarkBold & \XSolidBrush \\
RNGDet++ \cite{rngdetplusplus2023} & RAL23 & \XSolidBrush & \CheckmarkBold & \XSolidBrush \\
SAM-Road \cite{hetang2024samroad} & CVPRW24 & \XSolidBrush & \CheckmarkBold & \XSolidBrush \\
SAM-Road++ \cite{yin2025samroadplusplus} & CVPR25 & \XSolidBrush & \CheckmarkBold & \XSolidBrush \\
MaGRoad \cite{guan2025beyond} & CVPR26 & \XSolidBrush & \CheckmarkBold & \XSolidBrush \\
\midrule
PolyWorld \cite{zorzi2022polyworld} & CVPR22 & \CheckmarkBold & \CheckmarkBold & \XSolidBrush \\
TopDiG \cite{yang2023topdig} & CVPR23 & \CheckmarkBold & \CheckmarkBold & \XSolidBrush \\
Re:PolyWorld \cite{zorzi2023re} & ICCV23 & \CheckmarkBold & \CheckmarkBold & \XSolidBrush \\
UniVector \cite{yan2025univector} & FR26 & \CheckmarkBold & \CheckmarkBold & \XSolidBrush \\
\midrule
\rowcolor{blue!10}
VecLang (Ours) & -- & \CheckmarkBold & \CheckmarkBold & \CheckmarkBold \\
\bottomrule
\end{tabular}
}
\end{table}

Polygon-based methods are mainly designed for closed objects with clear boundaries, such as buildings, and usually recover object contours through mask regularization~\cite{girard2021polygonal,wei2023hisup} or coordinate regression~\cite{zhang2024p2pformer}. Early studies mainly focus on building outline extraction and have made significant progress in modeling the boundary structures of regular areal objects~\cite{wei2023hisup,zhang2024p2pformer,zhang2026vectorllm}. For example, Frame Field Learning~\cite{girard2021polygonal} introduces orientation fields to guide mask regularization, thereby improving the regularity and geometric accuracy of building contours. HiSup~\cite{wei2023hisup} jointly predicts corners, edges, and masks for stable polygon reconstruction, while P2PFormer~\cite{zhang2024p2pformer} formulates building outline extraction as a closed point sequence regression problem. Beyond building-centric vectorization, recent works attempt to extend polygon-based representations to larger remote sensing images and more diverse geographic objects. For instance, Li et al.~\cite{li2026full} explore full-scene vectorization for large-size imagery with objects such as roads and water bodies, and ACPV-Net~\cite{jiao2026acpvnet} further extends polygon-based vectorization to more land-cover categories, including vegetation, bare land, and water bodies. Nevertheless, polygon representations are inherently tied to closed structures, making them less suitable for network-like objects with explicit topological connections, such as road networks~\cite{yan2025univector,jiao2026acpvnet}.

Graph-based methods \cite{rngdetplusplus2023, hetang2024samroad, guan2025beyond} are more suitable for network-like objects such as road networks, as they represent vector objects as graphs of nodes and edges to explicitly model road centerlines and their connectivity . Existing studies \cite{hetang2024samroad, yin2025samroadplusplus, guan2025beyond} have shown strong performance in road network extraction. For example, Sat2Graph \cite{he2020sat2graph} encodes point locations and their connections into a tensor representation before reconstructing road networks through post-processing. RNGDet++ \cite{rngdetplusplus2023} progressively builds road networks by iteratively predicting node connections in local regions. SAM-Road \cite{hetang2024samroad} and SAM-Road++ \cite{yin2025samroadplusplus} introduce foundation segmentation models \cite{kirillov2023segment} to improve extraction robustness in complex scenes. However, graph representations usually lack explicit instance-level constraints, making it difficult to naturally describe object boundaries and ensure structural closure for closed objects such as buildings \cite{he2020sat2graph}.

Overall, as summarized in Table~\ref{tab:method_summary}, existing RSVM methods have achieved strong task-specific performance, but their category-specific representations make it difficult to uniformly model geometry, semantics, and topology across heterogeneous geospatial entities. In this work, we revisit RSVM from the perspective of representation and describe different types of entities with a shared structured language, enabling unified multiclass vector mapping.

\subsection{Remote Sensing Vision-Language Models}

In recent years, vision-language models (VLMs) have rapidly advanced remote sensing scene understanding, shifting the field from visual perception toward unified multimodal reasoning~\cite{baltrusaitis2019multimodal,zhang2024vlmsurvey}. Existing studies mainly fall into two lines: multimodal foundation models that learn joint representations~\cite{zeng2024x2vlm,guo2024skysense,zhang2025skysensev2}, and large-language-model-based methods that support language-driven tasks such as visual question answering, region captioning, object grounding, and open-vocabulary perception~\cite{zhang2024vlmsurvey,zhu2024openvocabulary,kuckreja2024geochat}.

For foundation models, SkySense~\cite{guo2024skysense} and SkySense V2~\cite{zhang2025skysensev2} enhance unified representation learning through large-scale multimodal pretraining, providing stronger features for downstream tasks. For vision-language understanding, GeoChat~\cite{kuckreja2024geochat}, GeoLLaVA-8K~\cite{wang2025geollava8k}, and ZoomSearch~\cite{zhou2025zoomsearch} move the field beyond closed task-specific learning toward more general language-driven modeling. Specifically, GeoChat~\cite{kuckreja2024geochat} supports visual question answering, region captioning, and referring object detection; GeoLLaVA-8K~\cite{wang2025geollava8k} improves the perception of ultra-high-resolution imagery with large-scale scenes and dense objects; and ZoomSearch~\cite{zhou2025zoomsearch} addresses multiscale object search and localization in large-format images. These studies show that VLMs can support image-level semantics, region-level localization, and language-guided fine-grained perception. Moreover, SegEarth-OV~\cite{li2025segearthov} and OpenRSD~\cite{huang2025openrsd} extend this line of work to open-vocabulary segmentation and open-prompt detection, showing that language can serve as a flexible semantic interface for cross-category generalization in complex remote sensing scenes.

Despite this progress, current remote sensing VLMs mainly produce text descriptions~\cite{hu2025rsgpt,pang2025vhm}, category labels~\cite{liu2024remoteclip}, bounding boxes~\cite{kuckreja2024geochat,huang2025openrsd}, or segmentation masks~\cite{li2025segearthov, li2025segearthov3, ni2025unigeoseg}, whereas RSVM requires explicit and executable structured maps that jointly encode the geometry, semantics, and topology of geospatial entities. As a result, the unified representation and generation of multiclass vector maps remain largely unexplored.

\subsection{Structured Text Generation}

With the rise of large language models and multimodal generative models, structured text generation has become increasingly important~\cite{xu2023multimodaltransformers,bie2025renaissance}. Recent work enables direct generation of JSON, code, and other parseable formats~\cite{beurerkellner2023lmql,scholak2021picard,li2022alphacode}, making outputs more readable, executable, and compatible with downstream systems. This capability has further been extended to explicit geometric representations, such as SVG images and CAD drawings~\cite{xing2025svgdreamerplusplus,wu2021deepcad,xu2022skexgen}.

For SVG image generation, models directly produce parseable representations such as paths, shapes, and hierarchical relations to generate vector graphics~\cite{carlier2020deepsvg,xing2025svgdreamerplusplus}. In CAD drawing generation and engineering design, models further generate programs, parameter sequences, or structured commands to describe 2D sketches, 3D parts, and engineering objects~\cite{wu2021deepcad,xu2022skexgen}. Similarly, some recent works in 3D vision use structured text or programmatic representations to organize spatial objects and their relations, such as scene graphs~\cite{chang2023scenegraphsurvey} and SpatialLM~\cite{mao2025spatiallm}. These studies show that structured text generation can express geometric information and complex structural constraints.

However, existing works mainly focus on graphic generation, engineering design, or spatial scene modeling, rather than explicit map generation for RSVM. Unlike SVG or CAD formats, vector maps require not only parseable syntax but also accurate encoding of geospatial geometry, semantics, and topology. Thus, structured generation for multiclass RSVM remains underexplored.

\begin{figure*}[!t]
    \centering
    \includegraphics[width=1.0\textwidth]{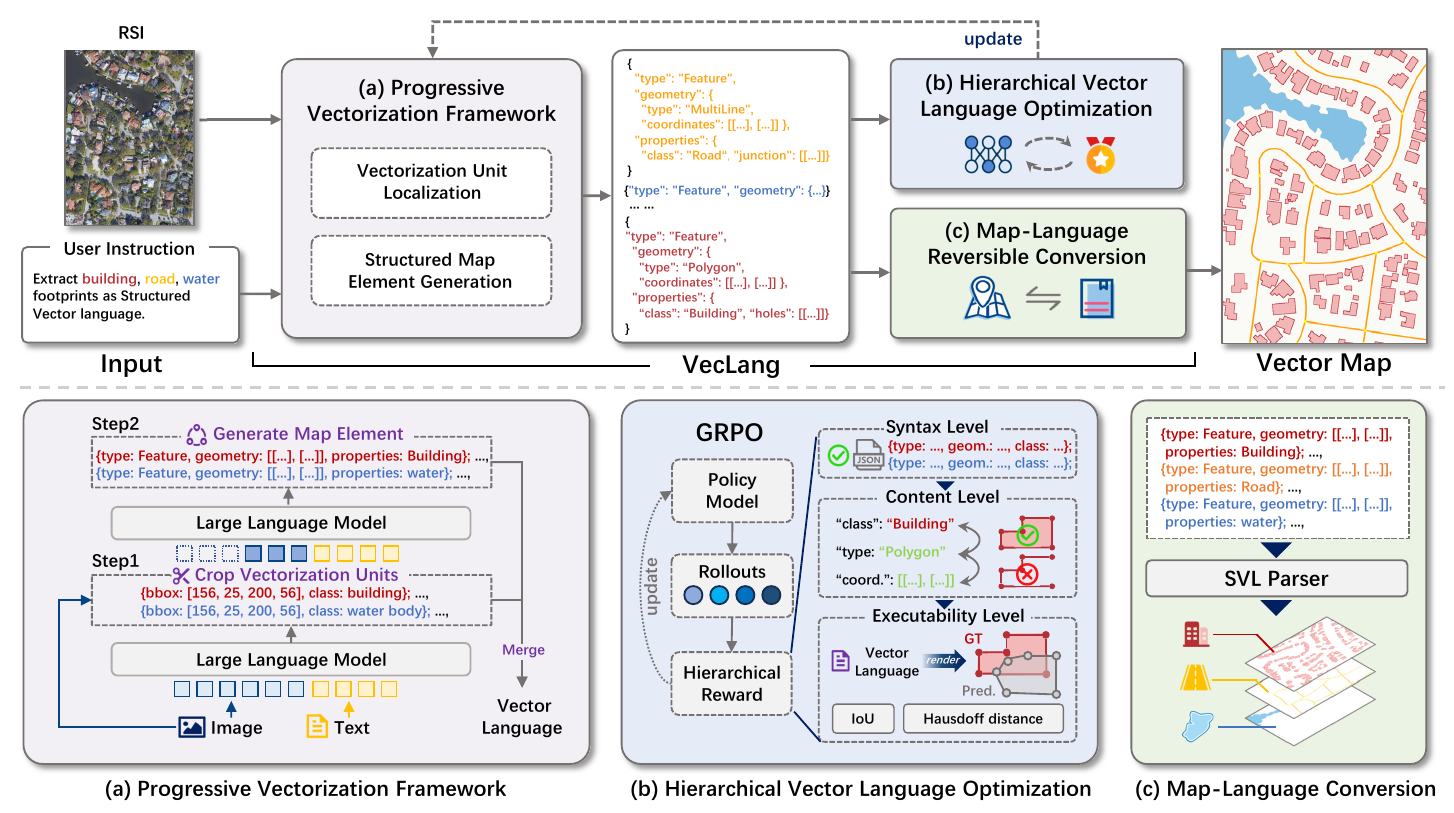}
    \caption{Overview of the VecLang framework. This pipeline consists of three parts: (a) the Progressive Vectorization Framework, which decomposes mapping into entity localization and language generation; (b) Hierarchical Vector Language Optimization, which improves vector language generation in terms of syntax, content, and executability; and (c) Map-Language Reversible Conversion, which constructs the Structured Vector Language representation and enables bidirectional conversion between maps and language.}
    \label{fig_method}
\end{figure*}

\section{Method}

\subsection{Overview}

Given a remote sensing image, we aim to generate a structured vector map containing geometry, semantics, and topology. Instead of using category-specific polygon or graph representations~\cite{hetang2024samroad,wei2023hisup}, we formulate remote sensing vector mapping as structured text generation, where heterogeneous map elements are encoded, generated, and optimized in a unified language space.

Formally, given an input image $I$, the target vector map $\mathcal{M}$ consists of a set of geospatial entities, where each entity contains its semantic class, geometric structure, and optional topological relations. Instead of predicting $\mathcal{M}$ in a category-specific format, VecLang converts it into a structured language sequence $Y=\{y_t\}_{t=1}^{T}$, where $y_t$ is the $t$-th token. The sequence $Y$ follows a predefined vector language grammar and can be parsed back into an executable vector map. The generation process is modeled as
\begin{equation}
    p_{\theta}(Y|I) = \prod_{t=1}^{T} p_{\theta}(y_t | I, y_{<t}),
\end{equation}
where $\theta$ denotes the parameters of the vision-language model.

Fig.~\ref{fig_method} illustrates the overall framework of VecLang, which contains three main components. First, we design Structured Vector Language (SVL), a GeoJSON-like representation with reversible map-language conversion for encoding geometry, semantics, and topology. Second, we propose a Progressive Vectorization Framework (PVF), which first localizes vectorization units and then generates structured map elements for each unit. Third, we introduce Hierarchical Vector Language Optimization (HVLO), which improves the generated vector language from three levels: syntax, content, and executability.

\subsection{Map-Language Reversible Conversion}

\subsubsection{Vector Structured Language (SVL)}

The representation of vector maps is the foundation of unified remote sensing vector mapping. To describe different categories of geospatial entities in a common form, we propose SVL, a GeoJSON-like \cite{butler2016geojson} language that represents vector maps as structured text. SVL is designed around three shared map elements: geometry, semantics, and topology.

As shown in Fig.~\ref{fig_motivation}(a), different categories of geospatial entities may exhibit diverse shapes, but they can be described by a common set of map elements: semantics, geometry, and topology. Based on this observation, each entity in SVL is represented as
\begin{equation}
    e_i = \{\text{id}_i, \text{class}_i, \text{geometry}_i, \text{topology}_i\}.
\end{equation}
The field $\text{id}_i$ identifies an entity instance, $\text{class}_i$ indicates its semantic category, and $\text{geometry}_i$ records its vector structure. For different categories, the geometry field can represent polygons, polylines, or multi-line structures with compositional relations. The field $\text{topology}_i$ describes relations such as containment and intersection.

This design allows different map objects to be represented in the same language space. For example, a building can be described by a semantic class and a polygon geometry, with holes used to express more complex structures. A road can be represented by a semantic class and a polyline or multi-line geometry, with intersections used to reflect crossing relations. Although these objects have different spatial structures, their information is organized by the same set of structured fields. As a result, SVL provides a unified representation for both closed objects and network-like objects.

SVL has three advantages. First, it unifies heterogeneous map elements in a single textual format, which enables category-agnostic modeling. Second, its structured syntax makes the output parseable and executable, allowing generated text to be converted into vector maps. Third, its GeoJSON-like design is compatible with common GIS-style data formats \cite{butler2016geojson}, which facilitates rendering, storage, and downstream spatial analysis.

\subsubsection{Map-Language Conversion}

\begin{figure*}[!t]
    \centering
    \includegraphics[width=1.0\textwidth]{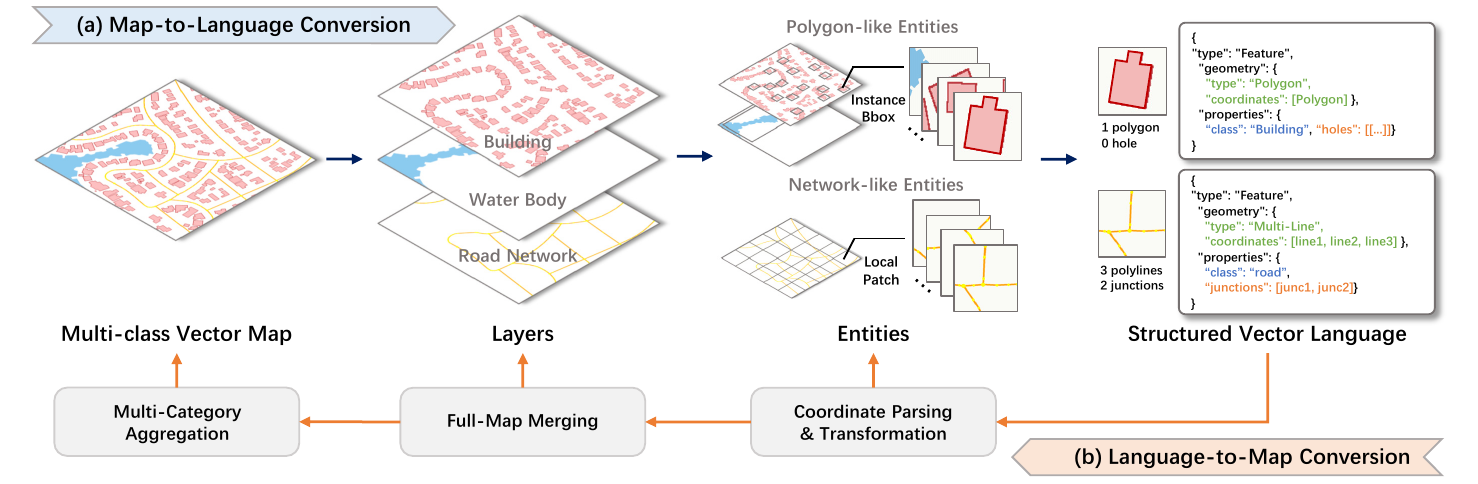}
    \caption{Map-Language Reversible Conversion. (a) Map-to-language conversion follows a top-down process, converting polygonal and network-like entities into unified Structured Vector Language. (b) Language-to-map conversion follows a bottom-up process, parsing entity-level Structured Vector Language into coordinates, then merging them into a complete multiclass map.}
    \label{fig_map-language}
\end{figure*}

To build a unified representation for training and inference, we convert vector map annotations into SVL sequences. However, this conversion is non-trivial because different geospatial entities have distinct structural forms \cite{yan2025univector}, as illustrated in Fig.~\ref{fig_map-language}. Polygonal objects are naturally defined by closed instance boundaries \cite{girard2021polygonal}, while road networks are continuous structures whose instance-level partitioning is ambiguous and difficult to standardize \cite{he2020sat2graph}.

To handle this structural heterogeneity, we adopt a structure-aware vectorization strategy, as shown in Fig.~\ref{fig_map-language}. The basic idea is to define vectorization units according to the structural characteristics of each category, and then describe every unit with the same SVL fields, including semantics, geometry, and topology. For polygonal categories, such as buildings and water bodies, each annotated instance naturally serves as a vectorization unit. Its semantic category is encoded as a class token, its geometry is represented by ordered boundary vertices, and its topology records holes when inner rings exist. For roads, instead of forcing the continuous network into independent instances, we treat each local patch as a vectorization unit. The road unit is encoded with the road category, represented by a polyline or multi-line geometry, and associated with junctions that describe intersections between line segments. With this design, both polygonal and network-like annotations can be serialized into unified SVL sequences while preserving their category-specific structures.

Formally, given a ground-truth vector map $\mathcal{M}^{*}$, we convert it into an SVL sequence through
\begin{equation}
    Y^{*} = \Phi(\mathcal{M}^{*}),
\end{equation}
where $\Phi(\cdot)$ denotes the map-to-language conversion function. As shown in Fig.~\ref{fig_map-language}(a), $\mathcal{M}^{*}$ is first organized into vectorization units and then serialized according to the SVL grammar.

The inverse process parses a generated SVL sequence back into an executable vector map:
\begin{equation}
    \hat{\mathcal{M}} = \Psi(\hat{Y}),
\end{equation}
where $\Psi(\cdot)$ denotes the language-to-map parser. As illustrated in Fig.~\ref{fig_map-language}(b), this process consists of three steps: coordinate parsing and transformation, whole-map merging, and category aggregation. Through this bidirectional conversion, VecLang connects structured text generation with executable vector map prediction.

\subsection{Progressive Vectorization Framework}

\subsubsection{Vectorization Unit Localization}

\begin{figure}[!t]
    \centering
    \includegraphics[width=0.5\textwidth]{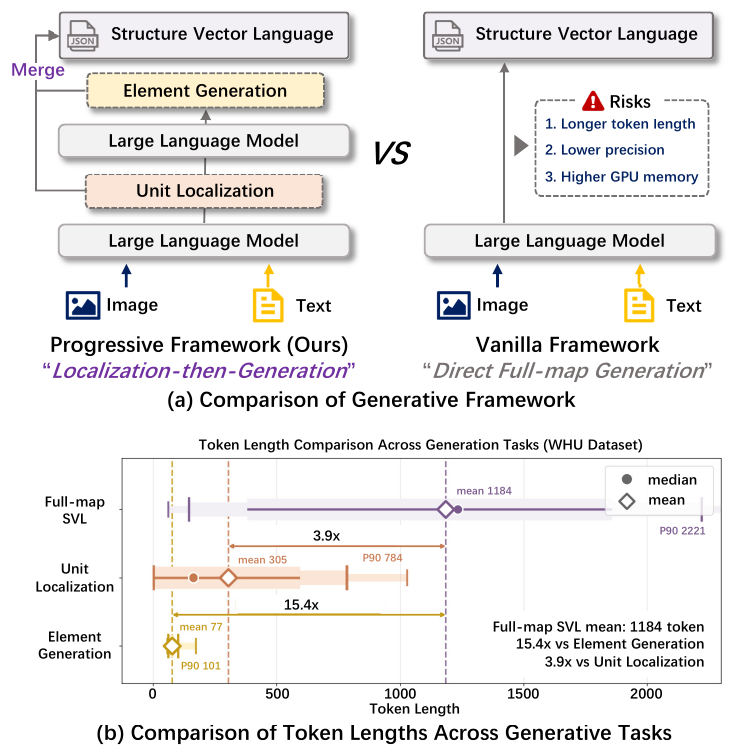}
    \caption{The effectiveness of the Progressive Generation Framework. (a) Direct full-text generation risks long sequences, lower accuracy, and high GPU memory use. (b) “Localization-then-generation” reduces the risk of each inference step.}
    \label{fig_framework}
\end{figure}

Vectorization unit localization provides visual grounding for structured map element generation. As shown in Fig.~\ref{fig_framework}, dense remote sensing scenes often contain numerous objects and complex structures, making direct full-map generation prone to long sequences, unstable reasoning, and high GPU memory consumption. PVF therefore adopts a localization-before-generation strategy, first identifying vectorization units and then predicting their structured map elements.

Instead of treating the entire image as a single generation target, PVF first decomposes the input image into a set of vectorization units. Each unit serves as a spatial anchor for later SVL generation, and its definition depends on the structural characteristics of the target category. For instance, a building can be treated as an object-level unit because it usually corresponds to a closed instance with clear boundaries \cite{girard2021polygonal}. In contrast, a road network is continuous and difficult to divide into independent instances under a universal rule \cite{he2020sat2graph}; therefore, we use a local region as the vectorization unit for roads, which may contain multiple polylines or multi-line structures.

In practice, we implement vectorization unit localization by adapting the vision-language model \cite{bai2025qwen3vl} to dense remote sensing prediction, as shown in Fig.~\ref{fig_framework}(1). Off-the-shelf VLMs often struggle to localize dense and small geospatial entities in high-resolution remote sensing images \cite{wang2025geollava8k}. To enhance this capability, we conduct post-training on our constructed benchmark, including supervised fine-tuning and reinforcement learning, so that the model can output bounding-box coordinates for specified categories in the image. For reinforcement learning, we adopt the Group Relative Policy Optimization (GRPO) \cite{shao2024deepseekmath} optimization algorithm with two rewards: an IoU reward \cite{huang2026rsgroundr1} that measures localization accuracy and a format reward \cite{zhou2025zoomsearch} that encourages valid and parseable coordinate outputs.

After localization, the predicted bounding boxes are used to define vectorization units. We then call a cropping tool to crop the corresponding image regions and resize them to a unified resolution. Each cropped unit is used as the localized visual input for subsequent instance-level structured map element generation. In this way, PVF converts dense whole-image prediction into localized unit-level generation, reducing sequence length and improving the stability of structured vector language generation.

\subsubsection{Structured Map Element Generation}

Structured map element generation aims to convert each localized vectorization unit into an SVL sequence. As shown in Fig.~\ref{fig_framework}(a), directly generating the whole vector map may lead to long output sequences, degraded accuracy, and high GPU memory consumption. Therefore, after vectorization unit localization, PVF avoids generating the whole map in a single pass and instead performs unit-level generation guided by category-specific textual instructions. This progressive process can be summarized as
\begin{equation}
    I \rightarrow \mathcal{U}=\{u_i\}_{i=1}^{K},
    \quad
    (u_i, q_i) \rightarrow \hat{Y}_i,
    \quad
    \{\hat{Y}_i\}_{i=1}^{K} \rightarrow \hat{\mathcal{M}},
\end{equation}
where $u_i$ is a localized vectorization unit, $q_i$ is the textual instruction specifying the target category and output format, $\hat{Y}_i$ is the generated unit-level SVL sequence, and $\hat{\mathcal{M}}$ is the final vector map. This formulation directly reflects the localization-before-reasoning strategy of PVF.

This task is challenging because the output must satisfy format, coordinate, and semantic constraints simultaneously. A valid SVL sequence should follow the predefined grammar, contain accurate coordinates, and assign correct categories. However, existing general-purpose VLMs are not specifically designed for dense remote sensing vector generation and often struggle to produce structured, coordinate-accurate, and executable outputs in complex scenes~\cite{wang2025geollava8k,zhou2025zoomsearch}. They may generate inconsistent formats, miss required fields, or produce coordinates with large spatial deviations, making the outputs difficult to parse into vector maps.

To address this issue, we adapt the VLM to structured vector language generation through task-specific post-training. Each vectorization unit provides localized visual evidence, while the textual instruction specifies the target category and output format. For closed objects, the generated sequence contains the object class, polygon boundary, and possible holes; for roads, it contains the road class, polyline or multi-line geometry, and junctions describing intersections between line segments.

The post-training data are derived from VecMap-Bench, where each vectorization unit is paired with an SVL annotation. We first conduct supervised fine-tuning to teach the model the SVL grammar and the mapping from localized visual evidence to structured map elements. We then apply reinforcement learning~\cite{zhang2026vectorllm,song2026simpleseg} to further improve format validity, coordinate accuracy, and map executability. With this design, structured map element generation becomes a localized and constrained text generation problem, reducing long-sequence burden and improving the stability of multiclass vector mapping.

\subsection{Hierarchical Vector Language Optimization}

\begin{figure}[!t]
    \centering
    \includegraphics[width=0.5\textwidth]{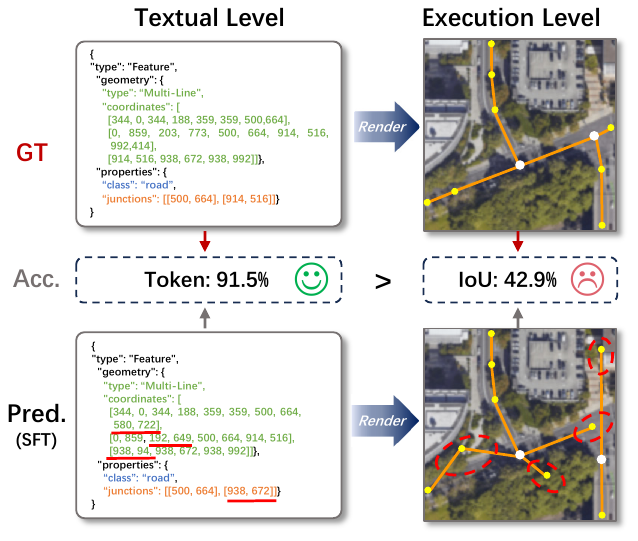}
    \caption{Challenge of Structured Vector Language Generation. Textual-level accuracy is not sufficient for Structured Vector Language generation, since small coordinate deviations can distort the executed map.}
    \label{fig_rl_motivation}
\end{figure}

Although supervised learning can teach the model to imitate SVL annotations, structured generation requires more than token-level accuracy when the output must be parsed and executed by downstream systems~\cite{wu2021deepcad,carlier2020deepsvg}. As shown in Fig.~\ref{fig_rl_motivation}, a sequence that is textually close to the target may still produce distorted vector maps after execution, since small coordinate deviations can significantly affect the final geometry. Therefore, a desirable prediction should have valid syntax, consistent content, and faithful execution. To this end, we introduce Hierarchical Vector Language Optimization, which uses reinforcement learning to optimize SVL generation at three levels: syntactic validity, content consistency, and execution fidelity.

Specifically, we optimize the model with GRPO~\cite{shao2024deepseekmath}. For each input prompt, the current policy samples a group of candidate SVL sequences $\{\hat{Y}_j\}_{j=1}^{G}$ and evaluates them with the hierarchical reward function. Let $R_j=R(\hat{Y}_j,Y^{*})$, and denote the mean and standard deviation of group rewards as $\bar{R}$ and $\sigma_R$, respectively. The relative advantage is computed as
\begin{equation}
    A_j = \frac{R_j-\bar{R}}{\sigma_R+\epsilon}.
\end{equation}
This group-relative normalization provides a stable learning signal without requiring an additional value model. The policy is then updated to favor candidates with higher relative rewards while constraining deviation from the reference policy:
\begin{equation}
    \mathcal{L}_{\text{GRPO}}
    =
    -\mathbb{E}_{j}
    \left[
    \min\left(r_j A_j, \tilde{r}_j A_j\right)
    \right]
    + \beta D_{\text{KL}}(\pi_{\theta}\|\pi_{\text{ref}}),
\end{equation}
where $r_j=\frac{\pi_{\theta}(\hat{Y}_j)}{\pi_{\text{old}}(\hat{Y}_j)}$ is the policy ratio, $\tilde{r}_j=\operatorname{clip}(r_j,1-\epsilon,1+\epsilon)$ is the clipped ratio, and $\pi_{\text{ref}}$ is the reference model.

The reward design is critical for reinforcement learning from feedback~\cite{huang2026rsgroundr1,shao2024deepseekmath}, because SVL outputs should be not only textually valid but also executable as vector maps. Thus, all rewards are defined over the generated sequence $\hat{Y}$ and the ground-truth sequence $Y^{*}$; when needed, they are parsed into vector entities $\hat{e}=\Psi(\hat{Y})$ and $e^{*}=\Psi(Y^{*})$. The final hierarchical reward is
\begin{equation}
    R = \lambda_{\text{syn}} R_{\text{syn}}
    + \lambda_{\text{con}} R_{\text{con}}
    + \lambda_{\text{exe}} R_{\text{exe}},
\end{equation}
where the three terms evaluate syntactic validity, content consistency, and execution fidelity, respectively.

\subsubsection{Syntax-Level Reward}

The syntax-level reward evaluates whether the generated sequence forms a valid vector language. In our implementation, this reward checks three aspects: JSON parsability, GeoJSON-like schema completeness, and field usability. Specifically, the generated text is first parsed into a feature object. We then check whether it contains required fields such as type, geometry, properties, geometry type, coordinates, and class label. For road objects, the junction field is also checked when it appears.

We define the syntax-level reward as
\begin{equation}
    R_{\text{syn}}(\hat{Y}, Y^{*}) =
    \mathbb{I}_{\text{json}}(\hat{Y})
    \left(
    \lambda_{\text{schema}} S_{\text{schema}}(\hat{Y}) +
    \lambda_{\text{field}} S_{\text{field}}(\hat{Y})
    \right),
\end{equation}
where $\mathbb{I}_{\text{json}}$ indicates whether the output can be parsed as JSON, $S_{\text{schema}}$ measures the completeness of the GeoJSON-like structure, and $S_{\text{field}}$ checks whether the key vector fields are usable for map construction. Although this reward is written with $Y^{*}$ for notational consistency, it only evaluates the validity of $\hat{Y}$. This gated design prevents invalid text from entering later content and geometric evaluation.

\subsubsection{Content-Level Reward}

The content-level reward evaluates whether the generated vector content is semantically and structurally consistent with the ground truth. Different from the syntax-level reward, this level does not only determine whether the output is parseable, but further examines whether the parsed content forms a reasonable vector object. Given $\hat{e}=\Psi(\hat{Y})$ and $e^{*}=\Psi(Y^{*})$, we define the content-level reward in a category-aware manner:
\begin{equation}
    R_{\text{con}}(\hat{Y}, Y^{*}) =
    \begin{cases}
        \mathcal{A}(S_{\text{type}}, V_{\text{poly}}, B_{\text{str}}),
        & c^{*} \in \mathcal{C}_{\text{closed}}, \\
        \mathcal{A}(S_{\text{type}}, V_{\text{line}}, B_{\text{str}}, S_{\text{junc}}),
        & c^{*} \in \mathcal{C}_{\text{road}},
    \end{cases}
\end{equation}
where $c^{*}$ is the class of $e^{*}$ and $\mathcal{A}(\cdot)$ denotes a bounded aggregation of sub-rewards. For closed objects, $S_{\text{type}}$ checks whether the predicted class and geometry type match the target, $V_{\text{poly}}$ evaluates whether the polygon is valid and non-degenerate, and $B_{\text{str}}$ measures whether the polygon has reasonable structural complexity by comparing the predicted and ground-truth vertex numbers. For roads, $S_{\text{type}}$ checks the road class and MultiLineString type, $V_{\text{line}}$ evaluates whether the line geometry is valid and non-empty, $B_{\text{str}}$ penalizes unnecessary fragmentation by comparing the predicted and ground-truth line component numbers, and $S_{\text{junc}}$ evaluates whether predicted junctions are parseable, attached to road lines, and spatially matched with ground-truth junctions. Empty road cases are explicitly handled, so negative samples without road geometries can also receive meaningful rewards.

\subsubsection{Executability-Level Reward}

The executability-level reward measures how well the parsed prediction matches the ground truth after being executed as vector geometry. Unlike the content-level reward, which focuses on structural plausibility, this level directly evaluates spatial fidelity and topology in the map space. Given $\hat{e}=\Psi(\hat{Y})$ and $e^{*}=\Psi(Y^{*})$, we define the executability-level reward in a category-aware manner:
\begin{equation}
    R_{\text{exe}}(\hat{Y}, Y^{*}) =
    \begin{cases}
        \mathcal{A}(\operatorname{IoU}, A_{c}),
        & c^{*} \in \mathcal{C}_{\text{closed}}, \\
        \mathcal{A}(\operatorname{IoU}_{\text{buf}}, A_{c}, S_{\text{conn}}),
        & c^{*} \in \mathcal{C}_{\text{road}},
    \end{cases}
\end{equation}
where $c^{*}$ is the class of $e^{*}$. For closed objects, this reward combines polygon IoU and boundary alignment. For roads, it combines buffered IoU, line alignment, and connectivity consistency \cite{heipke1997evaluation, he2020sat2graph}. The buffered IoU makes overlap evaluation stable for thin road structures, while standard polygon IoU is used for closed objects.

The alignment term is based on a normalized Hausdorff distance \cite{huttenlocher1993comparing}:
\begin{equation}
    A_{c}(\hat{g}, g^{*}) =
    \max\left(
    0,
    1 -
    \frac{
    d_{\mathrm{H}}(\Gamma_{c}(\hat{g}), \Gamma_{c}(g^{*}))
    }{
    \tau s(g^{*})
    }
    \right),
\end{equation}
where $d_{\mathrm{H}}$ denotes the Hausdorff distance \cite{huttenlocher1993comparing}, $s(g^{*})$ is the diagonal length of the ground-truth bounding box, and $\tau$ is a normalization factor. For closed objects, $\Gamma_{c}(\cdot)$ extracts polygon boundaries; for roads, it uses the line geometry itself. This term encourages accurate boundaries for polygons and precise centerline alignment for roads.

For road topology, the default lightweight connectivity score compares the number of connected components after buffering:
\begin{equation}
    S_{\text{conn}}(\hat{g}, g^{*}) =
    \frac{1}{1 + |\hat{n}_{\text{comp}} - n^{*}_{\text{comp}}|},
\end{equation}
where $\hat{n}_{\text{comp}}$ and $n^{*}_{\text{comp}}$ are the connected-component numbers of the predicted and ground-truth road geometries. The implementation also supports a graph-level mode that compares road nodes, edges, and connected components, but the lightweight form is more stable and efficient for RL optimization.

\section{Experiments}

\subsection{Experimental Setup}

\subsubsection{Benchmark Construction}

\begin{figure}[!t]
    \centering
    \includegraphics[width=0.5\textwidth]{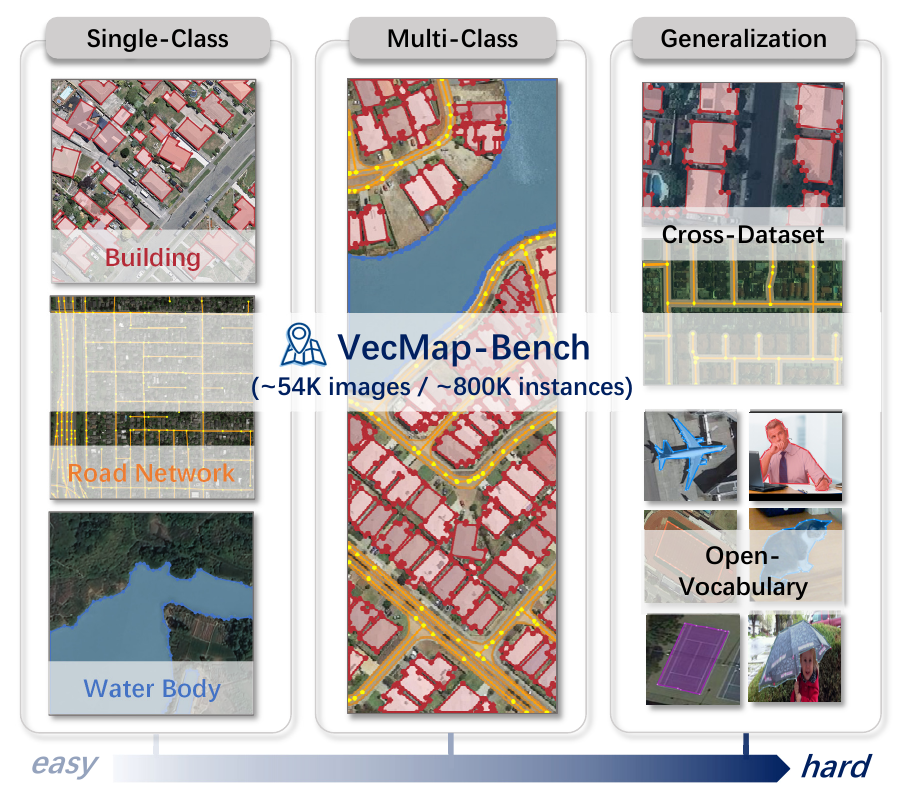}
    \caption{VecMap-Bench includes single-class vector mapping, multiclass vector mapping, and cross-dataset and open-vocabulary generalization tests, progressing from easy to difficult settings.}
    \label{fig_benchmark}
\end{figure}

To comprehensively evaluate unified remote sensing vector mapping, we construct \textbf{VecMap-Bench}, which covers single-class vector mapping, multiclass vector mapping, and generalization evaluation, as shown in Fig.~\ref{fig_benchmark}. VecMap-Bench contains about 54K images and 800K instances, spanning heterogeneous geospatial structures, including polygonal objects such as buildings and water bodies, and network-like objects such as roads.

Following the easy-to-difficult evaluation protocol illustrated in Fig.~\ref{fig_benchmark}, we organize VecMap-Bench into several settings. For single-class evaluation, we use WHU~\cite{ji2019whu} for building vectorization and CityScale~\cite{he2020sat2graph} for road network vectorization. For waterbody vectorization, we build Vec-WB from EvLab-SS~\cite{zhang2017evlabss}, GID~\cite{tong2020gid}, and WAQS~\cite{zhang2023aqsnet}. For multiclass evaluation, we construct a unified setting based on IRSAMap~\cite{meng2025irsamap}, where the model is required to jointly generate geometry, semantics, and topology for multiple categories. For generalization evaluation, we use CrowdAI~\cite{mohanty2020deep} and SpaceNet~\cite{vanetten2018spacenet} for cross-dataset transfer, and COCO~\cite{lin2014coco} and iSAID~\cite{zamir2019isaid} for open-vocabulary evaluation on natural and remote sensing scenes.

VecMap-Bench is constructed through a unified map-to-language pipeline. We first clean the original datasets by removing low-quality annotations, incomplete objects, and inconsistent image-label pairs. We then simplify annotated vertices to reduce redundant points while preserving the main geometric structure. Finally, all annotations are converted into Structured Vector Language, where each vectorization unit is represented with unified semantic, geometric, and topological fields. This process organizes heterogeneous annotations into a consistent structured language format for unified training, evaluation, and generalization analysis.

\subsubsection{Evaluation Metrics}

We evaluate vector mapping performance according to the structural type of target entities. For polygonal objects, such as buildings and water bodies, we report instance-level mAP~\cite{lin2014coco}, pixel-level IoU~\cite{wei2023hisup}, and vector-geometry metrics including C-IoU~\cite{zorzi2022polyworld} and PoLiS~\cite{avbelj2015metric}. Among them, mAP evaluates instance detection and matching quality, while IoU measures the overlap between predicted and ground-truth regions. C-IoU and PoLiS further reflect the quality of vectorized geometry, including geometric fidelity and vertex compactness.

For network-like objects such as roads, we follow previous works~\cite{he2020sat2graph} and adopt topology-level precision, recall, and F1-score~\cite{hetang2024samroad}, together with APLS~\cite{he2020sat2graph} for connectivity evaluation. These metrics measure whether the predicted road network correctly recovers centerline geometry and preserves graph connectivity. For multiclass vector mapping, we report per-category results to evaluate the unified modeling ability across heterogeneous entity structures. For generalization settings, we use the same metrics on unseen datasets or unseen categories to assess cross-dataset and open-vocabulary transferability.

\begin{table*}[!t]
\centering
\caption{Single-class vector mapping performance on building, road, and water body datasets. 
For mAP, IoU, C-IoU, precision, recall, F1, and APLS, higher values indicate better performance; for PoLiS, lower values are better. The best and second-best results in each column are highlighted with dark and light blue backgrounds, respectively. $^\dagger$ denotes results reproduced by ourselves.}
\label{tab:single_class_vector_mapping}
\resizebox{\textwidth}{!}{
\begin{tabular}{lcc|cccc|cccc|cccc}
\toprule
\multirow{2}{*}{Method} & \multirow{2}{*}{Pub.} & \multirow{2}{*}{Training} 
& \multicolumn{4}{c|}{WHU (Building)} 
& \multicolumn{4}{c|}{Cityscales (Road)} 
& \multicolumn{4}{c}{Vector-WB (Water Body)} \\
\cmidrule(lr){4-7} \cmidrule(lr){8-11} \cmidrule(lr){12-15}
& & 
& mAP$\uparrow$ & IoU$\uparrow$ & C-IoU$\uparrow$ & PoLiS$\downarrow$
& Prec.$\uparrow$ & Rec.$\uparrow$ & F1$\uparrow$ & APLS$\uparrow$
& mAP$\uparrow$ & IoU$\uparrow$ & C-IoU$\uparrow$ & PoLiS$\downarrow$ \\
\midrule

\rowcolor{gray!15}
\multicolumn{15}{l}{\textit{Vision-Language Models}} \\
Gemini-3.1-Flash-Lite \cite{deepmind2026gemini31flashlite} & -- & \XSolidBrush 
& 3.10 & 33.75 & 22.68 & 12.76
& 67.04 & 37.03 & 47.71 & 43.07
& 0.78 & 47.45 & 29.84 & 28.64 \\
Qwen3.5-Plus \cite{alibaba2026qwen35plus} & -- & \XSolidBrush 
& 5.32 & 47.90 & 27.20 & 11.58
& 61.24 & 12.35 & 20.40 & 3.72
& 0.76 & 38.64 & 20.66 & 34.85 \\
VHM \cite{pang2025vhm} & AAAI25 & \XSolidBrush 
& 0.00 & 3.04 & 1.21 & 21.72
& 83.09 & 0.92 & 0.18 & 0.30
& 0.00 & 7.49 & 2.85 & 51.05 \\
GeoLLaVA-8K \cite{wang2025geollava8k} & NeurIPS25 & \XSolidBrush 
& 0.10 & 11.17 & 1.26 & 79.76
& 0.00 & 0.00 & 0.00 & 0.00
& 1.11 & 35.97 & 14.84 & 77.20 \\
ZoomSearch \cite{zhou2025zoomsearch} & CVPR26 & \XSolidBrush 
& 0.00 & 6.45 & 4.24 & 18.27
& 0.00 & 0.00 & 0.00 & 0.00
& 0.00 & 8.26 & 5.28 & 39.15 \\

\midrule
\rowcolor{gray!15}
\multicolumn{15}{l}{\textit{Building Vectorization Models}} \\
FFL \cite{girard2021polygonal} & CVPR21 & \CheckmarkBold 
& -- & 80.86 & 46.09 & 2.13
& -- & -- & -- & --
& -- & -- & -- & -- \\
HiSup$^\dagger$ \cite{wei2023hisup} & ISPRS23 & \CheckmarkBold 
& 69.83 & 90.12 & \second{84.36} & 1.54
& -- & -- & -- & --
& -- & -- & -- & -- \\
P2PFormer$^\dagger$ \cite{zhang2024p2pformer} & TGRS24 & \CheckmarkBold 
& 69.55 & 75.07 & 68.73 & 1.55
& -- & -- & -- & --
& -- & -- & -- & -- \\
HoliTracer \cite{wang2025holitracer} & ICCV25 & \CheckmarkBold 
& 61.07 & \second{91.60} & 82.30 & 3.63
& -- & -- & -- & --
& -- & -- & -- & -- \\
VectorLLM \cite{zhang2026vectorllm} & ISPRS26 & \CheckmarkBold 
& 78.30 & -- & -- & --
& -- & -- & -- & --
& -- & -- & -- & -- \\
ACPV-Net$^\dagger$ \cite{jiao2026acpvnet} & CVPR26 & \CheckmarkBold 
& \second{79.39} & 91.13 & 77.90 & \second{1.51}
& -- & -- & -- & --
& -- & -- & -- & -- \\

\midrule
\rowcolor{gray!15}
\multicolumn{15}{l}{\textit{Road Vectorization Models}} \\
Sat2Graph \cite{he2020sat2graph} & ECCV20 & \CheckmarkBold 
& -- & -- & -- & --
& 80.70 & 72.28 & 76.26 & 63.14
& -- & -- & -- & -- \\
RNGDet++ \cite{rngdetplusplus2023} & RAL23 & \CheckmarkBold 
& -- & -- & -- & --
& 85.65 & 72.58 & \second{78.44} & 67.76
& -- & -- & -- & -- \\
SAM-Road \cite{hetang2024samroad} & CVPRW24 & \CheckmarkBold 
& -- & -- & -- & --
& \best{90.47} & 67.69 & 77.23 & \second{68.37}
& -- & -- & -- & -- \\
SAM-Road++ \cite{yin2025samroadplusplus} & CVPR25 & \CheckmarkBold 
& -- & -- & -- & --
& \second{88.39} & \second{73.39} & \best{80.01} & 68.34
& -- & -- & -- & -- \\
MaGRoad \cite{guan2025beyond} & CVPR26 & \CheckmarkBold 
& -- & -- & -- & --
& 84.46 & 72.66 & 78.11 & \best{71.27}
& -- & -- & -- & -- \\

\midrule
\rowcolor{gray!15}
\multicolumn{15}{l}{\textit{Water Body Vectorization Models}} \\
HiSup$^\dagger$ \cite{wei2023hisup} & ISPRS23 & \CheckmarkBold 
& -- & -- & -- & --
& -- & -- & -- & --
& 61.39 & 83.96 & \second{66.04} & 5.44 \\
P2PFormer$^\dagger$ \cite{zhang2024p2pformer} & TGRS24 & \CheckmarkBold 
& -- & -- & -- & --
& -- & -- & -- & --
& 45.52 & 54.45 & 12.94 & 7.92 \\
ACPV-Net$^\dagger$ \cite{jiao2026acpvnet} & CVPR26 & \CheckmarkBold 
& -- & -- & -- & --
& -- & -- & -- & --
& \second{63.55} & \best{87.20} & 58.86 & \second{5.01} \\

\midrule
\rowcolor{gray!15}
\multicolumn{15}{l}{\textit{Unified Vectorization Models}} \\
VecLang (Ours) & -- & \CheckmarkBold 
& \best{88.96} & \best{92.22} & \best{92.01} & \best{0.85}
& 72.21 & \best{75.96} & 73.95 & 64.05
& \best{64.82} & \second{86.77} & \best{76.88} & \best{4.96} \\

\bottomrule
\end{tabular}
}
\end{table*}

\begin{figure*}[!t]
    \centering
    \includegraphics[width=\textwidth]{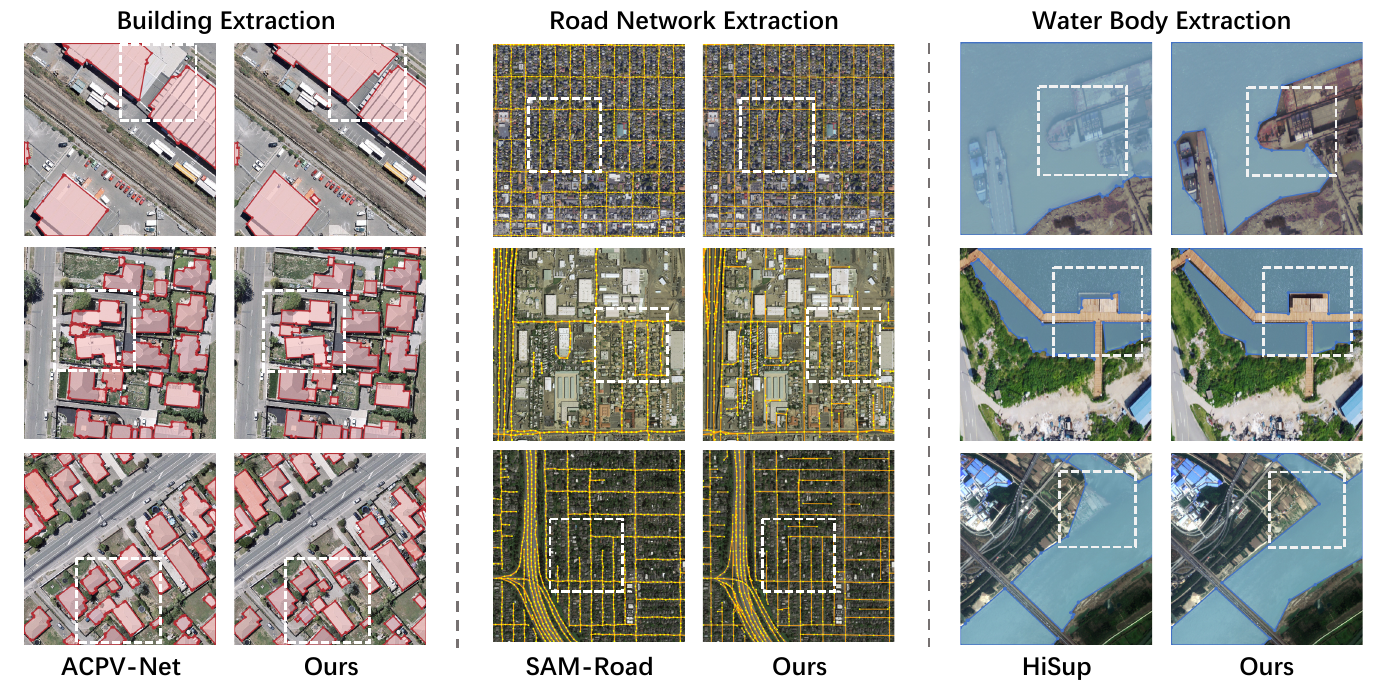}
    \caption{Visual comparison of VecLang on three remote sensing vector mapping tasks \cite{jiao2026acpvnet, hetang2024samroad}.}
    \label{fig_single_class_viz}
\end{figure*}

\subsubsection{Implementation Details}

We implement VecLang based on Qwen3-VL-4B~\cite{bai2025qwen3vl}, which serves as the base vision-language model. The model is trained in two stages. In the first stage, we perform LoRA-based supervised fine-tuning (SFT)~\cite{hu2022lora} using paired remote sensing images and Structured Vector Language annotations, enabling the model to learn the mapping from visual content to structured vector language. In the second stage, we apply GRPO-based hierarchical vector language reinforcement learning to further improve syntax validity, content consistency, and execution fidelity.

For SFT, we use LoRA fine-tuning with rank 16 and $\alpha=32$, and train the model for 3 epochs with a learning rate of $5\times10^{-5}$. The maximum sequence length is set to 2300, the per-device batch size is 2, and the gradient accumulation step is 8. For GRPO, we initialize the model from the SFT checkpoint and continue LoRA-based optimization with rank 8 and $\alpha=32$. The learning rate is set to $5\times10^{-6}$, the per-device batch size is 4, and the model is trained for 0.3 epochs with bfloat16 precision. We use four generations for each prompt and optimize the model with the proposed hierarchical rewards, whose weights for syntax, content, and execution are set to 0.1, 0.2, and 0.7, respectively.

\subsection{Unified Vector Mapping Performance}

\subsubsection{Single-Class Vector Mapping}

We first evaluate VecLang on single-class vector mapping tasks, including building extraction on WHU, road network extraction on Cityscales, and waterbody vectorization on Vector-WB. These tasks cover both polygonal objects and network-like structures, enabling comparison with category-specific vectorization methods under different geometric forms.

As shown in Table~\ref{tab:single_class_vector_mapping}, VecLang achieves the best overall performance on polygonal objects. On WHU, it obtains 88.96 mAP, 92.22 IoU, 92.01 C-IoU, and 0.85 PoLiS, outperforming existing building vectorization methods across all metrics. On Vector-WB, it also achieves the best mAP, C-IoU, and PoLiS, indicating its ability to generate accurate and compact polygonal geometry. For road vectorization, VecLang achieves the highest recall of 75.96, showing strong coverage of road structures. Its precision is slightly lower than specialized road methods, mainly because the generative nature of large vision-language models may introduce a few hallucinated road segments. Nevertheless, VecLang remains competitive on road extraction while providing a unified solution for both closed objects and network-like structures.

The qualitative results in Fig.~\ref{fig_single_class_viz} further support these findings. VecLang produces regular building outlines, compact waterbody polygons, and continuous road networks within the same language-based framework. For roads, although a few redundant segments may appear, the generated networks contain cleaner and more compact nodes, leading to visually simpler topology. These results show that VecLang can handle diverse single-class vector mapping tasks without relying on category-specific representations.

\begin{table*}[!t]
\centering
\caption{Multi-class vector mapping performance on building, road, and water body categories. 
For mAP, IoU, C-IoU, precision, recall, F1, and APLS, higher values indicate better performance; for PoLiS, lower values are better. * indicates that we vectorize the mask using the Douglas-Peucker algorithm.}
\label{tab:multi_class_vector_mapping}
\resizebox{\textwidth}{!}{
\begin{tabular}{lcc|cccc|cccc|cccc}
\toprule
\multirow{2}{*}{Method} & \multirow{2}{*}{Pub.} & \multirow{2}{*}{Vector}
& \multicolumn{4}{c|}{Building}
& \multicolumn{4}{c|}{Road}
& \multicolumn{4}{c}{Water Body} \\
\cmidrule(lr){4-7} \cmidrule(lr){8-11} \cmidrule(lr){12-15}
& & 
& mAP$\uparrow$ & IoU$\uparrow$ & C-IoU$\uparrow$ & PoLiS$\downarrow$
& Prec.$\uparrow$ & Rec.$\uparrow$ & F1$\uparrow$ & APLS$\uparrow$
& mAP$\uparrow$ & IoU$\uparrow$ & C-IoU$\uparrow$ & PoLiS$\downarrow$ \\
\midrule

Mask R-CNN* \cite{he2017mask} & ICCV17 & \XSolidBrush
& 27.35 & 64.19 & 54.85 & 2.46
& 84.09 & 19.81 & 30.48 & 7.83
& 28.55 & 81.51 & 74.36 & 5.02 \\

Mask2Former* \cite{cheng2022mask2former} & CVPR22 & \XSolidBrush
& 32.13 & 68.15 & 48.35 & 2.47
& 81.03 & 41.47 & 53.37 & 16.74
& 32.79 & 83.39 & 76.35 & \best{2.99} \\

UniVector \cite{yan2025univector} & FR26 & \CheckmarkBold
& \second{34.52} & \second{74.26} & \second{62.12} & \second{2.20}
& 82.50 & 48.66 & 61.24 & \second{35.22}
& \second{34.24} & 82.15 & 78.33 & 4.25 \\

ACPV-Net \cite{jiao2026acpvnet} & CVPR26 & \CheckmarkBold
& 10.93 & 71.70 & 60.17 & 2.94
& \best{88.12} & \second{54.70} & \second{66.07} & 34.77
& 21.21 & \second{83.92} & \second{78.59} & 3.92 \\

\midrule
VecLang (Ours) & -- & \CheckmarkBold
& \best{71.04} & \best{89.26} & \best{82.41} & \best{0.89}
& \second{85.12} & \best{71.04} & \best{77.00} & \best{38.45}
& \best{50.65} & \best{93.63} & \best{82.03} & \second{3.50} \\

\bottomrule
\end{tabular}
}
\end{table*}

\begin{figure*}[!t]
    \centering
    \includegraphics[width=1.0\textwidth]{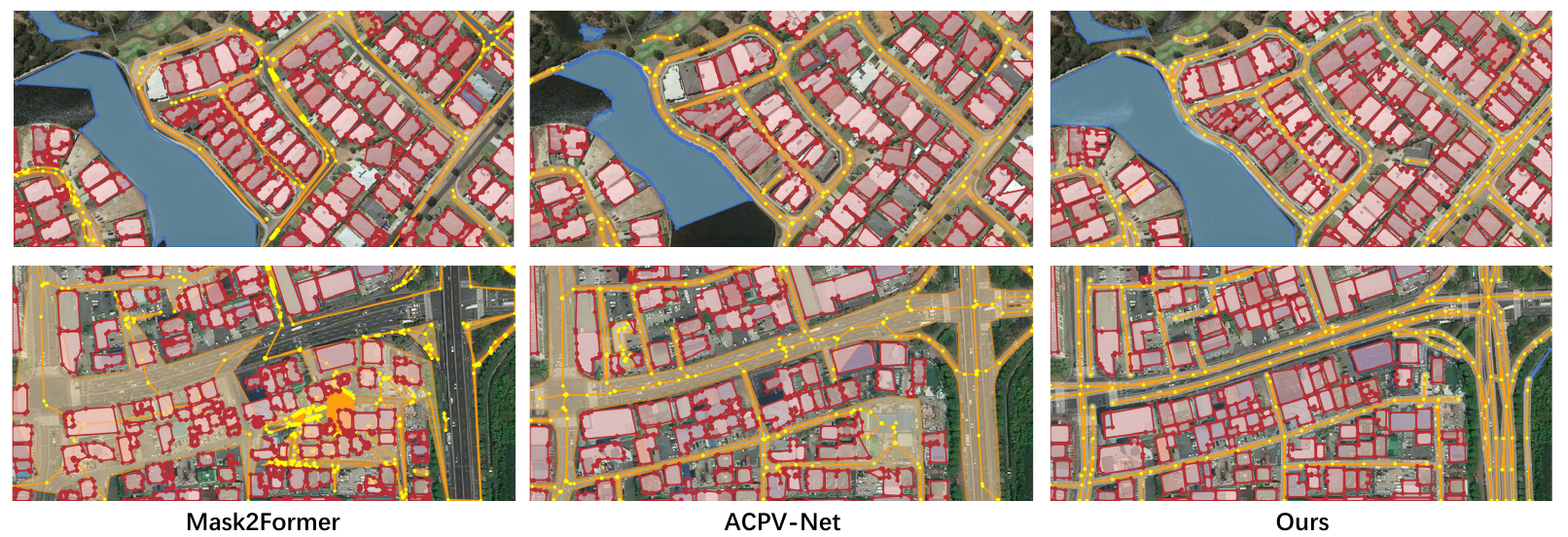}
    \caption{Visual comparison of VecLang on the multiclass vector mapping task \cite{cheng2022mask2former,jiao2026acpvnet}.}
    \label{fig_multi_class_viz}
\end{figure*}

\subsubsection{Multi-Class Vector Mapping}

We further evaluate VecLang in a multiclass vector mapping setting, where buildings, roads, and water bodies are predicted within a single framework. This setting is more challenging than single-class evaluation, as the model must jointly distinguish semantic categories and generate category-dependent geometry and topology.

As shown in Table~\ref{tab:multi_class_vector_mapping}, VecLang achieves the best overall performance across heterogeneous categories. For buildings, it reaches 71.04 mAP, 89.26 IoU, 82.41 C-IoU, and 0.89 PoLiS, substantially outperforming existing methods. For roads, VecLang obtains the best recall, F1, and APLS, indicating stronger ability to recover complete and connected road structures. For water bodies, it achieves the best mAP, IoU, and C-IoU, with the second-best PoLiS. These results show that VecLang can jointly model polygonal and network-like entities without category-specific output heads or separate vector representations.

The visual results in Fig.~\ref{fig_multi_class_viz} further support this advantage. VecLang produces accurate building boundaries, coherent road networks, and complete waterbody polygons in the same scene. In contrast, existing multiclass methods based on polygon extraction can recover closed objects to some extent, but their polygon representations struggle to express complex road connections, leading to fragmented or topologically inaccurate road networks. VecLang better preserves both instance-level shapes and network connectivity, confirming the effectiveness of representing multiclass vector maps as structured language.

\subsection{Generalization Evaluation}

\subsubsection{Cross-Dataset Generalization}

To evaluate cross-dataset generalization, we train models on source datasets and test them on target datasets with different geographic regions, imaging conditions, and annotation styles. Specifically, we evaluate building transfer from WHU to CrowdAI and road transfer from Cityscales to SpaceNet, examining whether the learned vector representation can generalize beyond the source distribution.

As shown in Table~\ref{tab:cross_dataset_generalization}, VecLang is robust under dataset shifts. For building transfer, it achieves the best results across all metrics, with 17.81 mAP, 53.05 IoU, 36.76 C-IoU, and 3.50 PoLiS, while specialized building methods degrade significantly on CrowdAI. For road transfer, VecLang obtains the best precision, recall, and F1-score, reaching 72.03, 29.15, and 37.60, respectively. Although SAM-Road achieves higher APLS, VecLang better preserves road completeness and topology-level matching under cross-dataset shifts.

The visual results in Fig.~\ref{fig_generalization}(a) further confirm this trend. VecLang produces more complete building outlines and road structures on unseen datasets, whereas task-specific methods often suffer from missing objects, fragmented geometry, or degraded topology. These results indicate that the shared Structured Vector Language helps the model reuse geometric and topological priors across datasets, improving transferability for remote sensing vector mapping.

% Cross-dataset generalization
\begin{table*}[!t]
\centering
\caption{Cross-dataset generalization performance. Models are trained on the source dataset and evaluated on the target dataset. For mAP, IoU, C-IoU, precision, recall, F1, and APLS, higher values indicate better performance; for PoLiS, lower values are better.}
\label{tab:cross_dataset_generalization}
\resizebox{\textwidth}{!}{
\begin{tabular}{lcc|cccc|cccc}
\toprule
\multirow{2}{*}{Method} 
& \multirow{2}{*}{Train}
& \multirow{2}{*}{Multi-class}
& \multicolumn{4}{c|}{WHU$\rightarrow$CrowdAI}
& \multicolumn{4}{c}{Cityscales$\rightarrow$SpaceNet} \\
\cmidrule(lr){4-7} \cmidrule(lr){8-11}
& & 
& mAP$\uparrow$ & IoU$\uparrow$ & C-IoU$\uparrow$ & PoLiS$\downarrow$
& Prec.$\uparrow$ & Rec.$\uparrow$ & F1$\uparrow$ & APLS$\uparrow$ \\
\midrule

\rowcolor{gray!15}
\multicolumn{11}{l}{\textit{Building Extraction Methods}} \\
HiSup \cite{wei2023hisup} & WHU & \XSolidBrush
& 0.73 & 1.41 & 0.70 & 17.73
& -- & -- & -- & -- \\
P2PFormer \cite{zhang2024p2pformer} & WHU & \XSolidBrush
& 1.24 & 20.47 & 10.57 & \second{6.74}
& -- & -- & -- & -- \\
ACPV-Net \cite{jiao2026acpvnet} & WHU & \XSolidBrush
& 0.19 & 6.36 & 4.49 & 12.46
& -- & -- & -- & -- \\

\midrule
\rowcolor{gray!15}
\multicolumn{11}{l}{\textit{Road Extraction Methods}} \\
RNGDet++ \cite{rngdetplusplus2023} & Cityscales & \XSolidBrush
& -- & -- & -- & --
& 14.93 & 1.56 & 2.61 & 7.12 \\
SAM-Road \cite{hetang2024samroad} & Cityscales & \XSolidBrush
& -- & -- & -- & --
& 33.50 & \second{18.10} & \second{21.49} & \best{69.33} \\
MaGRoad \cite{guan2025beyond} & Cityscales & \XSolidBrush
& -- & -- & -- & --
& 27.11 & 12.62 & 15.29 & \second{54.32} \\

\midrule
\rowcolor{gray!15}
\multicolumn{11}{l}{\textit{Vision-Language Models}} \\
Gemini-3.1-Flash-Lite \cite{deepmind2026gemini31flashlite} & -- & \CheckmarkBold
& \second{6.87} & \second{50.73} & \second{31.92} & 6.87
& \second{36.25} & 4.54 & 8.07 & 6.85 \\
Qwen3.5-Plus \cite{alibaba2026qwen35plus} & -- & \CheckmarkBold
& 5.74 & 49.78 & 26.89 & 8.07
& 29.13 & 2.32 & 3.99 & 3.52 \\
GeoLLaVA-8K \cite{wang2025geollava8k} & -- & \CheckmarkBold
& 0.00 & 0.00 & 0.00 & --
& 0.00 & 0.00 & 0.00 & 0.00 \\

\midrule
Qwen3-VL-4B (Base) \cite{bai2025qwen3vl} & -- & \CheckmarkBold
& 2.32 & 22.45 & 14.33 & 9.55
& 13.25 & 1.77 & 2.58 & 2.02 \\
VecLang (Ours) & WHU, Cityscales & \CheckmarkBold
& \best{17.81} & \best{53.05} & \best{36.76} & \best{3.50}
& \best{72.03} & \best{29.15} & \best{37.60} & 32.97 \\

\bottomrule
\end{tabular}
}
\end{table*}

% Open-vocabulary generalization on remote sensing categories
\begin{table}[!t]
\centering
\caption{Open-vocabulary generalization results on remote sensing categories. Higher values indicate better performance. ``Base VLM'' denotes Qwen3-VL-4B.}
\label{tab:open_vocabulary_remote_sensing}
\resizebox{\columnwidth}{!}{
\begin{tabular}{lc|ccccc}
\toprule
\multirow{2}{*}{Method} & \multirow{2}{*}{Vector}
& \multicolumn{5}{c}{Remote Sensing Category} \\
\cmidrule(lr){3-7}
& & Plane & Soccer & Pool & Tennis & Mean \\
\midrule
SegEarth-OV \cite{li2025segearthov} & \XSolidBrush
& 34.79 & \second{54.08} & 22.68 & 24.62 & 34.04 \\
SAM3 \cite{carion2025sam3} & \XSolidBrush
& \second{55.18} & 11.98 & 18.34 & 39.49 & 31.25 \\
UniGeoSeg \cite{ni2025unigeoseg} & \XSolidBrush
& 36.81 & \best{78.75} & 0.16 & 46.60 & 40.58 \\
SegEarth-OV3 \cite{li2025segearthov3} & \XSolidBrush
& 53.30 & 16.44 & \second{45.87} & \second{48.11} & \second{40.93} \\
\midrule
Base VLM \cite{bai2025qwen3vl} & \CheckmarkBold
& 15.65 & 20.58 & 19.52 & 17.62 & 18.34 \\
VecLang (Ours) & \CheckmarkBold
& \best{62.54} & 52.89 & \best{81.00} & \best{92.00} & \best{78.22} \\
\bottomrule
\end{tabular}
}
\end{table}

\subsubsection{Open-Vocabulary Generalization}

To examine the flexibility of VecLang, we evaluate open-vocabulary generalization on unseen remote sensing categories using textual prompts. This setting tests whether the model can extend structured vector generation beyond the categories used during training.

As shown in Table~\ref{tab:open_vocabulary_remote_sensing}, VecLang performs strongly on unseen remote sensing categories, achieving the best mean score of 78.22 and outperforming open-vocabulary segmentation methods \cite{li2025segearthov, li2025segearthov3, carion2025sam3},. It obtains the best results on plane, swimming pool, and tennis court, and remains competitive on soccer ball field. Compared with the base Qwen3-VL-4B model, VecLang improves the mean score from 18.34 to 78.22, indicating that structured vector-language training greatly enhances open-category vector mapping ability.

\begin{figure}[!t]
    \centering
    \includegraphics[width=0.5\textwidth]{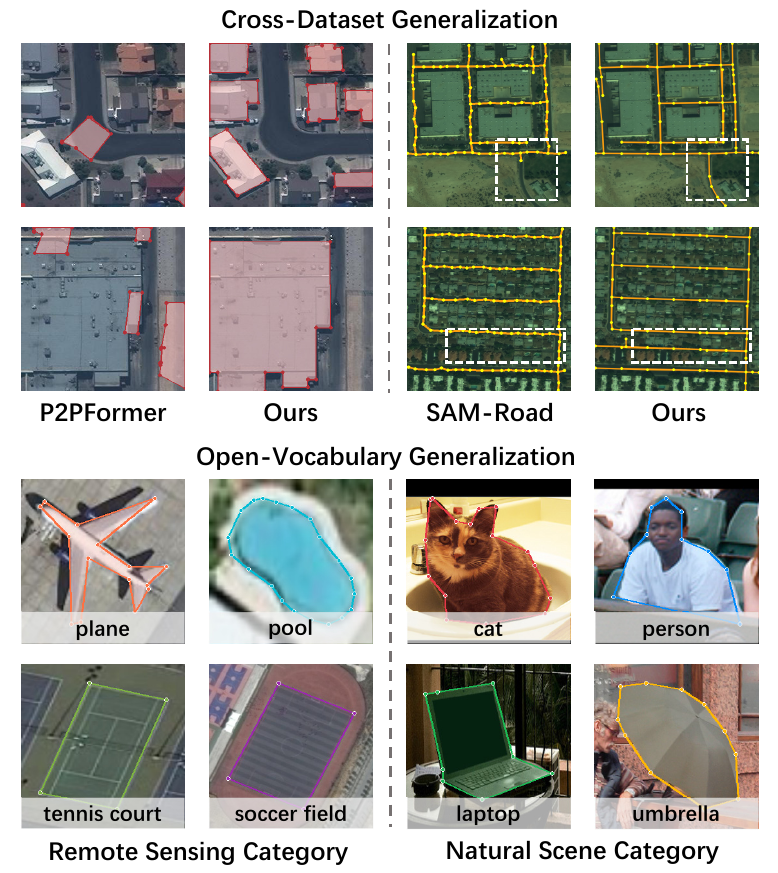}
    \caption{Performance demonstration of VecLang on cross-dataset and open-vocabulary generalization tasks. The results show that our post-training enables the base VLM to acquire strong open-vocabulary vector mapping ability.}
    \label{fig_generalization}
\end{figure}

The qualitative results in Fig.~\ref{fig_generalization} further show that VecLang can localize and vectorize unseen remote sensing objects with clear structural boundaries. For compact objects such as planes and swimming pools, VecLang produces complete and regular vector contours. For larger areal objects such as soccer fields and tennis courts, it preserves the overall shape and spatial extent while generating executable polygonal outputs. These results suggest that the learned structured language representation can transfer to unseen remote sensing categories with similar geometric patterns.

In addition to the remote sensing evaluation, we provide qualitative results on natural scene categories in Fig.~\ref{fig_generalization}. Although these categories are outside the remote sensing domain, VecLang can still generate meaningful structured vector outputs from textual prompts. This suggests that fine-tuning with structured vector representations can endow VLMs with certain open-vocabulary segmentation and vectorization abilities, demonstrating the flexibility of the proposed language-based formulation.

\subsubsection{Large-Scale Scene Vectorization}

To further examine the scalability and generalization ability of VecLang, we conduct large-scale inference on high-resolution remote sensing scenes outside the standard benchmark. Since dense vector annotations are unavailable for these scenes, we focus on qualitative visualization. This setting assesses whether VecLang can generalize beyond benchmark images and generate coherent, executable vector maps over large spatial extents with dense and heterogeneous geospatial entities. Due to space limitations, the main paper presents only a representative slice from a large-scale scene, while complete large-scale visualization results are provided in the supplementary material and on our project page.\footnote{\url{https://github.com/yyyyll0ss/VecLang}}

\begin{figure*}[!t]
    \centering
    \includegraphics[width=1.0\textwidth]{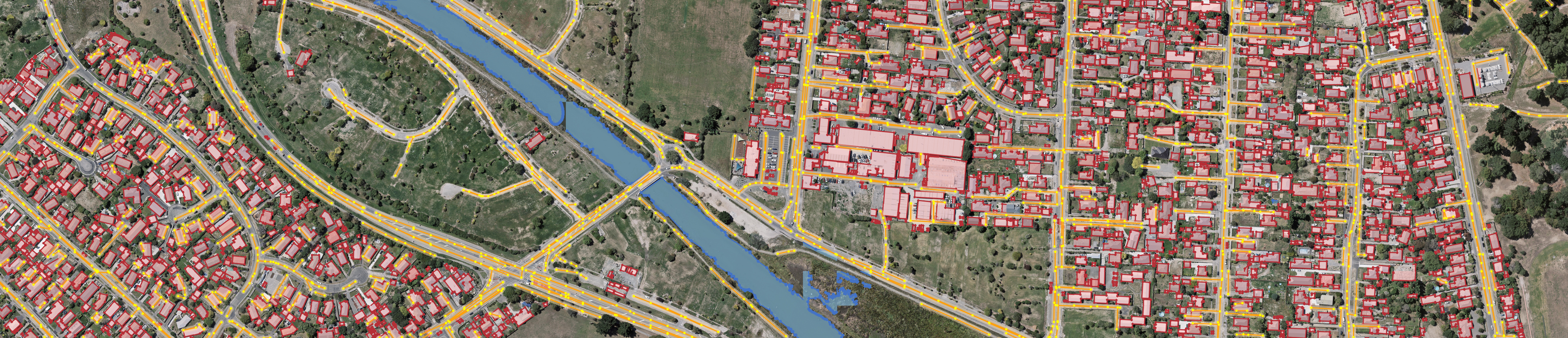}
    \caption{Large-scale vectorization results on images outside the benchmark. The displayed result shows a representative slice of a complete high-resolution scene-level vector map; full large-scale visualization results are provided in the supplementary material and on our project page. VecLang generalizes to high-resolution scenes and produces executable scene-level vector maps.}
    \label{fig_large_scene}
\end{figure*}

As shown in Fig.~\ref{fig_large_scene}, VecLang generates structured vector maps by progressively localizing vectorization units, producing unit-level SVL sequences, and merging them into complete scene-level maps. Although the figure shows only a cropped region of the full large-scale output, the visualization demonstrates that VecLang preserves boundaries of closed entities and maintains continuity of network-like structures within dense and heterogeneous scenes. These results indicate good generalization ability on unseen large-scale remote sensing scenes. All vector maps are directly parsed from generated SVL outputs without manual correction.

\subsection{Ablation Studies and Analysis}

\subsubsection{Component Ablation}

We conduct component ablation studies to analyze the contribution of Structured Vector Language (SVL), Progressive Vectorization Framework (PVF), and Hierarchical Vector Language Reinforcement Learning (HVRL). As shown in Table~\ref{tab:component_ablation}, the base Qwen3-VL-4B model performs poorly on structured vector mapping, achieving only 18.36 on buildings and 13.20 on roads. Introducing SVL improves the scores to 50.24 and 35.88, showing that a structured language representation is essential for vector map generation. Adding PVF further boosts the performance to 90.30 and 68.12 by decomposing dense scenes into localized vectorization units. Finally, HVRL brings additional gains, reaching the best scores of 92.01 and 73.95. These results show that representation, progressive generation, and reward-based optimization all contribute to the final performance.

% Component ablation study of VecLang
\begin{table}[!t]
\centering
\caption{Component ablation study of VecLang.}
\label{tab:component_ablation}
\resizebox{0.90\columnwidth}{!}{
\begin{tabular}{lccc|cc}
\toprule
Method & SVL & PVF & HVRL & Building & Road \\
\midrule
Qwen3-VL-4B & \XSolidBrush & \XSolidBrush & \XSolidBrush & 18.36 & 13.20 \\
+ SVL & \CheckmarkBold & \XSolidBrush & \XSolidBrush & 50.24 & 35.88 \\
+ PVF & \CheckmarkBold & \CheckmarkBold & \XSolidBrush & 90.30 & 68.12 \\
+ HVRL & \CheckmarkBold & \CheckmarkBold & \CheckmarkBold & \best{92.01} & \best{73.95} \\
\bottomrule
\end{tabular}
}
\end{table}

% Representation ablation study.
\begin{table}[!t]
\centering
\caption{Representation ablation study. Parse, Cls., Geo., and Topo. are reported in percentage, while Len. denotes the average token length.}
\label{tab:representation_ablation}
\resizebox{0.90\columnwidth}{!}{
\begin{tabular}{lccccc}
\toprule
Rep. & Parse & Cls. & Geo. & Topo. & Len. \\
\midrule
Coord. Seq. & \best{99.88} & 91.40 & 88.20 & 53.66 & \best{126} \\
Free Text & 73.25 & 86.70 & 67.02 & 38.55 & 214 \\
Plain JSON & 99.20 & \second{96.42} & \second{90.55} & \second{54.24} & 168 \\
SVL & \second{99.75} & \best{96.80} & \best{92.01} & \best{64.05} & \second{142} \\
\bottomrule
\end{tabular}
}
\end{table}

\begin{table}[!t]
\centering
\caption{Framework ablation study. Parse and mAP are reported in percentage. Time/Inst. and Time/Image denote the average inference time, and GPU Mem. denotes the peak GPU memory usage.}
\label{tab:framework_ablation}
\resizebox{\columnwidth}{!}{
\begin{tabular}{lccccc}
\toprule
Framework & Parse & mAP & Time/Inst. & Time/Image & GPU Mem. \\
\midrule
ACPV-Net \cite{jiao2026acpvnet} & -- & \second{79.39} & -- & \second{3.81s} & \best{763M} \\
Vanilla Framework & \second{74.53} & 35.89 & -- & 5.85s & 5.5G \\
PVF (Ours) & \best{99.75} & \best{88.96} & 0.16s & \best{3.39s} & \second{1.25G} \\
\bottomrule
\end{tabular}
}
\end{table}

% Reward ablation study
\begin{table}[!t]
\centering
\caption{Reward ablation study of hierarchical vector-language reinforcement learning.}
\label{tab:reward_ablation}
\resizebox{\columnwidth}{!}{
\begin{tabular}{ccc|cc}
\toprule
Syntax-Level & Content-Level & Execution-Level & Building & Road \\
\midrule
\XSolidBrush & \XSolidBrush & \XSolidBrush & 90.30 & 68.12 \\
\CheckmarkBold & \XSolidBrush & \XSolidBrush & 85.23 & 64.33 \\
\CheckmarkBold & \CheckmarkBold & \XSolidBrush & 90.68 & 68.75 \\
\XSolidBrush & \CheckmarkBold & \CheckmarkBold & 78.51 & 48.66 \\
\CheckmarkBold & \XSolidBrush & \CheckmarkBold & \second{90.92} & \second{71.33} \\
\CheckmarkBold & \CheckmarkBold & \CheckmarkBold & \best{92.01} & \best{73.95} \\
\bottomrule
\end{tabular}
}
\end{table}

\begin{figure}[!t]
    \centering
    \includegraphics[width=0.5\textwidth]{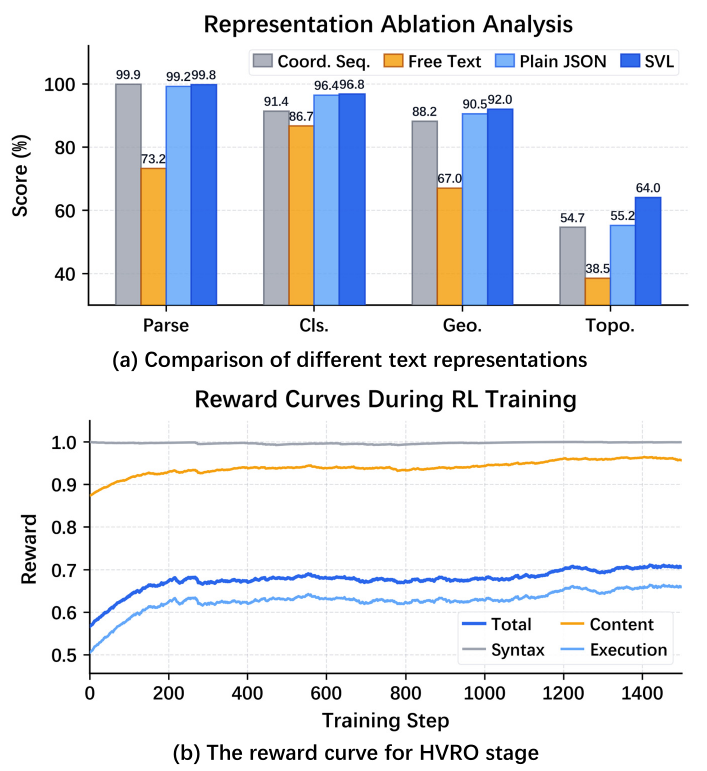}
    \caption{(a) Performance comparison of different text representations. (b) Training curves of different rewards during the reinforcement learning stage.}
    \label{fig_ablation}
\end{figure}

\subsubsection{Representation Ablation}

We study the effect of different vector representations, including coordinate sequences, free-form text, plain JSON, and the proposed SVL. As shown in Table~\ref{tab:representation_ablation}, coordinate sequences achieve the highest parse rate and shortest length due to their simple format, but their topology score is limited to 53.66 because explicit structural fields are missing. Free text performs worst on most metrics, indicating that natural language descriptions are difficult to parse and execute reliably. Plain JSON improves structural consistency, but its topology score remains 54.24 since relations are still encoded with ad-hoc fields. In contrast, SVL achieves the best class, geometry, and topology scores of 96.80, 92.01, and 64.05, respectively, while keeping a compact token length of 142. Fig.~\ref{fig_ablation}(a) further shows that syntactic simplicity alone is insufficient for vector map representation. Overall, SVL provides a better balance between parsability, geometric fidelity, topology modeling, and sequence efficiency.

\subsubsection{Framework Ablation}

We evaluate the effectiveness of the Progressive Vectorization Framework (PVF) in Table~\ref{tab:framework_ablation}. Compared with vanilla full-image generation, PVF improves Parse from 74.53 to 99.75 and mAP from 35.89 to 88.96, showing that localization-before-generation greatly improves the stability and accuracy of structured vector generation. PVF also achieves the fastest image-level inference time of 3.39s with only 1.25G GPU memory. Compared with the non-VLM lightweight model ACPV-Net, PVF does not suffer from slower inference; instead, it achieves better speed and substantially higher mAP, while keeping memory usage competitive. These results demonstrate that PVF provides a better balance between accuracy, efficiency, and resource consumption.

\subsubsection{Reward Ablation}

We analyze Hierarchical Vector Language Optimization by using different combinations of syntax-level, content-level, and execution-level rewards. As shown in Table~\ref{tab:reward_ablation}, using all three rewards achieves the best performance, with 92.01 on buildings and 73.95 on roads. Compared with the setting without reinforcement learning, the full reward design improves the road score from 68.12 to 73.95, showing its benefit for topology-sensitive road vectorization.

The ablation results indicate that the three rewards are complementary. The syntax-level reward alone does not improve performance, since format validity cannot ensure correct geometry or topology. Combining syntax and content rewards brings limited gains, while adding the execution-level reward gives stronger improvement, especially on roads. Removing the syntax-level reward also causes a clear drop, indicating that valid structured language remains necessary for reliable execution. Fig.~\ref{fig_ablation}(b) further shows that the total reward increases steadily during training, with syntax, content, and execution rewards providing complementary optimization signals. These results confirm the importance of jointly optimizing syntactic validity, content consistency, and execution fidelity.

\section{Conclusion}

In this paper, we presented VecLang, a unified framework for representing and generating heterogeneous vector maps in remote sensing vector mapping. VecLang moves beyond separate polygon- or graph-based representations by formulating vector mapping as structured language generation, where geometry, semantics, and topology are organized in a shared language space. With a GeoJSON-like Structured Vector Language and a progressive vision-language framework, diverse geospatial entities can be encoded, generated, and converted into executable vector maps. Hierarchical Vector Language Optimization further improves syntactic validity, content consistency, and execution fidelity. Extensive experiments on VecMap-Bench across single-class, multiclass, cross-dataset, and open-vocabulary settings show that VecLang effectively unifies different vector mapping tasks and achieves strong accuracy and generalization. These results suggest that representing vector maps as language provides a scalable paradigm for unified remote sensing vector mapping.

In future work, we plan to extend VecLang to more challenging scenarios, including open-category vector mapping, temporal map updating, and remote sensing VLMs with stronger spatial reasoning and multimodal geospatial understanding. We hope this work encourages further exploration of language-based paradigms for structured remote sensing understanding.

% use section* for acknowledgment
\ifCLASSOPTIONcompsoc
% The Computer Society usually uses the plural form
\section*{Acknowledgments}
\else
% regular IEEE prefers the singular form
\section*{Acknowledgment}
\fi This work was supported in part by the National Natural Science Foundation of China under the Original Exploration Program 62550100, the Distinguished Young Scholars Program 62425109, and Grant U22B2014, and in part by the Science and Technology Plan Project Fund of Hunan Province under Grant 2022RC3064.

\bibliography{ref}
\bibliographystyle{IEEEtran}

% that's all folks
\end{document}